\documentclass{article}

\usepackage{arxiv}
\usepackage[utf8]{inputenc} 
\usepackage[T1]{fontenc}    
\usepackage{hyperref}       
\usepackage{url}            
\usepackage{booktabs}       
\usepackage{amsfonts}       
\usepackage{nicefrac}       
\usepackage{microtype}      
\usepackage{lipsum}		    
\usepackage{graphicx}
\usepackage[square,numbers,sort&compress]{natbib} 
\usepackage{doi}
\usepackage{kotex} 
\usepackage{amsmath} 
\usepackage{multirow}
\usepackage[left]{lineno} 
\usepackage{xcolor} 
\usepackage{tabularx} 
\usepackage{array} 

\begin{document}


\title{Generative AI-Driven Parametric Design Optimization Framework for Vehicle Closure Systems}

\date{}                     


\author{
  \textbf{Dongeon Lee}$^{a,\dagger}$,
  \textbf{Yurim Lee}$^{a,\dagger}$,
  \textbf{Jiyong Kim}$^{a}$,
  \textbf{Sanghyeon Kim}$^{b}$, \\
  \textbf{Sanghyun Lee}$^{c}$,
  \textbf{Seungyeon Shin}$^{c}$,
  \textbf{Jeahong Kim}$^{c}$,
  \textbf{Namwoo Kang}$^{d,e,}$\thanks{Corresponding author: nwkang@kaist.ac.kr} \\[1.25em]
  \textit{$^{a}$Cho Chun Shik Graduate School of Mobility, Korea Advanced Institute of Science and Technology,} \\
  \textit{Daejeon, 34051, Republic of Korea} \\
  \textit{$^{b}$Division of Future Vehicle, Korea Advanced Institute of Science and Technology,} \\
  \textit{Daejeon, 34141, Republic of Korea} \\
  \textit{$^{c}$Hyundai Motor Company, Hwaseong, 18280, Republic of Korea} \\
  \textit{$^{d}$Department of Mechanical Engineering, Korea Advanced Institute of Science and Technology,} \\
  \textit{Daejeon, 34141, Republic of Korea} \\
  \textit{$^{e}$Narnia Labs, Daejeon, 34051, Republic of Korea} \\[1.25em]
  \textit{$^{\dagger}$These authors contributed equally to this work.}
}

\renewcommand{\headeright}{}
\renewcommand{\undertitle}{}
\renewcommand{\shorttitle}{}

\setlength{\parindent}{15pt} 

\hypersetup{
pdftitle={Generative AI-Driven Parametric Design Optimization Framework for Vehicle Closure Systems},
pdfsubject={industrial application of generative AI, parametric CAD/CAE automation, surrogate modeling, generative modeling, and multiobjective design optimization},
pdfauthor={Dongeon Lee, Yurim Lee, Jiyong Kim, Sanghyeon Kim, Sanghyun Lee, Seungyeon Shin, Jeahong Kim, Namwoo Kang},
pdfkeywords={Generative AI, Parametric design, CAD/CAE automation, Surrogate modeling, Design optimization, Vehicle closure system},
}

\maketitle

\begin{abstract} 
This study presents a generative AI-driven parametric design optimization framework that accelerates design exploration by connecting performance-oriented design generation with editable computer-aided design (CAD) and downstream computer-aided engineering (CAE) analysis for industrial vehicle closure systems. The framework is demonstrated on an automotive tailgate assembly characterized by complex geometry, strongly coupled assembly-level components, and competing performance and CAD/CAE feasibility requirements. A production-oriented CATIA model is restructured into a reference-based parametric template with a 13-dimensional design representation, and an automated CAD/CAE workflow generates parametric models, performs structural analysis, and constructs a design-performance database of mass, maximum displacement, and maximum von Mises stress. Gaussian process regression surrogate models predict performance, while diffusion- and flow matching-based generative models generate design candidates. The design problem minimizes mass and maximum displacement, with maximum von Mises stress treated as an engineering feasibility constraint. Engineering objectives are incorporated through performance-guided generation and performance-aware training, with sensitivity to guidance scale and loss weight examined. Generative search is compared with NSGA-II under matched evaluation budgets using a common surrogate model. Selected Pareto-region candidates are reconstructed in CATIA and re-evaluated through native CAE, with all verified candidates successfully regenerated and analyzed. The knee-point design reduced mass and maximum displacement by 7.1\% and 29.4\%, respectively, relative to the baseline, while the extreme verified solutions achieved maximum reductions of 10.3\% in mass and 37.3\% in maximum displacement. The results demonstrate that generative AI can support performance-oriented multi-objective design exploration while preserving compatibility with editable industrial CAD and native engineering verification.
\end{abstract}

\vspace{1.00em}


\clearpage

\section{Introduction}
\label{sec:introduction}
Artificial intelligence (AI) is increasingly reshaping engineering design by reorganizing the roles of humans and machines and enabling broader design exploration, faster iteration, and tighter feedback across the design process \cite{menold2026greater}. More broadly, AI-driven design can involve complementary roles as predictor, generator, and optimizer, which can be integrated to support design evaluation and optimization across the engineering design process \cite{yan2026ai}. In this context, generative AI is increasingly being adopted in engineering design and optimization, especially for conceptual design activities, to generate diverse design alternatives, accelerate the design process, and facilitate the search for high-performing designs \cite{oh2019deep,regenwetter2022deep,kang2025generative}. However, performance-oriented generation alone is insufficient for industrial product development, because the validity of generated designs also depends on how geometric dependencies and parametric relationships are represented in the underlying computer-aided design (CAD) model. For direct integration into industrial product-development workflows, the design representation should therefore preserve the geometric and parametric relationships required for robust CAD regeneration and downstream computer-aided engineering (CAE) analysis. Establishing such a representation provides a practical bridge between early-stage generative design exploration and editable CAD/CAE-based detailed engineering.

Achieving this integration remains challenging in industrial design workflows. Many data-driven generative approaches employ design representations of varying dimensionality, ranging from 1D parametric variables and 2D images to 3D geometric representations such as voxels, point clouds, and meshes, to facilitate geometric learning and design exploration \cite{regenwetter2022deep,kang2025generative}. However, such representations do not necessarily preserve the feature history, parent--child references, assembly relationships, and modeling conventions embedded in native parametric CAD models. Moreover, the underlying modeling strategy can strongly influence whether automated geometry variations remain valid throughout design exploration \cite{aranburu2022does}. Consequently, a parameterized design vector that satisfies the prescribed design-variable bounds may still fail during CAD regeneration, while a geometrically valid CAD model may subsequently fail during meshing, structural analysis, or constraint verification. For industrial generative model-based design optimization, feasibility with respect to prescribed design-variable bounds should therefore be distinguished from feasibility within the native CAD/CAE workflow.

Vehicle closure systems provide a representative industrial setting for this challenge. Doors, hoods, trunk lids, tailgates, and related body structures comprise interacting components such as body panels, reinforcements, glazing, hinges, latching mechanisms, and sealing components, whose geometric and performance requirements are strongly coupled \cite{malen2020fundamentals,lee2023multidisciplinary,lee2025multidisciplinary,lee2025design}. Consequently, changes in key dimensions or exterior design requirements can propagate through dependent CAD features, requiring simultaneous consideration of geometric, mass, structural, and manufacturing requirements. Recent studies on automotive hoods have also established large-scale 3D geometry--performance datasets to support representation learning, surrogate modeling, and data-driven design optimization under structural and multiphysics settings \cite{wollstadt2022carhoods10k,sharma2026autohood3d}. For tailgate-specific design optimization, recent work has addressed this challenge by combining parametric rib design, finite element analysis (FEA), Gaussian process regression (GPR) surrogate modeling, and Bayesian optimization (BO) under manufacturing constraints \cite{lee2025bayesian}. While BO provides a data-efficient strategy for identifying high-performing designs through sequential candidate evaluation, its search proceeds iteratively within a predefined parametric design space. Generative models offer an alternative design-search approach by learning the underlying design distribution and generating multiple design alternatives once trained. However, translating this capability into industrial product development requires integration with editable parametric CAD representations, automated engineering performance evaluation, and native CAD/CAE feasibility verification.

Beyond workflow integration, a further challenge is to generate candidates that are not only diverse but also consistently directed toward high-performing regions of the design space. Prior studies have incorporated performance information into generative models either by conditioning generation on prescribed performance targets \cite{bucher2023performance} or by using surrogate predictions or gradients to guide the generative process toward desired engineering performance \cite{maze2023diffusion,yang2026physics}. These studies motivate two complementary mechanisms for incorporating engineering performance into generative design: guiding the generation process toward high-performing regions and embedding performance awareness during model training. In this study, these roles are addressed through performance-guided generation using surrogate guidance and performance awareness during training using a performance-aware loss. However, overly strong surrogate guidance may drive generated candidates toward sparsely sampled or out-of-distribution regions, where surrogate uncertainty and prediction error can increase \cite{yang2026physics}. Achieving effective and stable generation therefore requires evaluating the effectiveness and sensitivity of both performance guidance and performance-aware training, while promising candidates should ultimately be reconstructed and re-evaluated in the original CAD/CAE environment.

Accordingly, this study presents a generative AI-driven parametric design optimization framework for vehicle closure systems and demonstrates its application to an automotive tailgate assembly in an industrial product development context. First, a production-oriented CATIA model is restructured into a reference-based parametric representation, and an automated CAD/CAE workflow is established to generate engineering data while recording CAD and CAE failures. Second, engineering performance is incorporated into generative models, including diffusion and flow matching, through performance-guided generation using surrogate guidance and performance-aware training using a performance-aware loss, with the effectiveness and sensitivity of both mechanisms systematically evaluated. Third, generative search is compared with the non-dominated sorting genetic algorithm II (NSGA-II) under matched evaluation budgets using a common surrogate model to characterize budget-dependent search behavior. Finally, selected Pareto-region candidates are reconstructed and re-evaluated in the original CAD/CAE environment to assess native CAD/CAE feasibility and surrogate prediction accuracy. Together, these elements establish an industrial generative design optimization workflow integrating editable parametric CAD, CAD/CAE automation, performance-oriented generative modeling, multi-objective optimization, and native CAD/CAE verification. Notably, its application to a production-derived tailgate model from Hyundai Motor Company demonstrates the ability to handle complex geometric dependencies and assess structural feasibility, highlighting its practical value for industrial engineering design and optimization.

The remainder of this paper is organized as follows. Section~\ref{sec:related_work} reviews related studies on generative AI for engineering design, AI-driven parametric CAD model generation, and surrogate-assisted design optimization. Section~\ref{sec:research_framework} presents the proposed generative AI-driven parametric design optimization framework. Sections~\ref{sec:data_construction} and \ref{sec:model_development_design_optimization} describe training data construction through CAD/CAE automation and AI model development and design optimization, respectively. Section~\ref{sec:case_study} presents the industrial case study, including the experimental results and discussion. Section~\ref{sec:conclusions_future_work} concludes the paper and discusses directions for future work.

\section{Related work}
\label{sec:related_work}

\subsection{Generative AI for engineering design and optimization}
\label{subsec:generative_design_optimization}
Generative design refers broadly to computational approaches that create and explore design alternatives under designer- or engineer-defined requirements \cite{shea2005towards}. Earlier performance-driven approaches combined parametric representations, engineering analysis, and optimization to explore feasible design alternatives, whereas deep generative models introduced the ability to learn distributions of existing designs and generate new candidates from the learned design space \cite{oh2019deep,regenwetter2022deep}. This development has expanded the role of generative AI in engineering from design synthesis toward inverse design and performance-oriented design exploration \cite{kang2025generative}. More recently, generative modeling has increasingly adopted diffusion and flow-matching approaches. Diffusion models generate samples through a learned reverse denoising process \cite{ho2020denoising,song2020denoising,song2020score}, whereas flow matching learns a continuous vector field that transports samples from a simple source distribution to the target data distribution \cite{lipman2022flow,tong2023improving}. Beyond reproducing the training distribution, these models can be conditioned or guided using external performance information, making them attractive for engineering design problems in which candidate generation must be directed toward desired performance regions.

Performance incorporation has therefore become an important direction in generative engineering design. Conditional generative models can produce designs according to prescribed performance targets \cite{bucher2023performance}, while surrogate predictions or gradients can be incorporated into the generative process to promote improved engineering performance \cite{maze2023diffusion,yang2026physics}. These studies demonstrate that generative models can support performance-oriented design exploration rather than merely reproduce existing designs. Beyond generative modeling itself, prior work has also integrated generative design with automated CAD/CAE workflows by automatically generating 3D CAD designs, performing CAE analysis, and evaluating engineering performance using learned predictive models \cite{yoo2021integrating}. However, performance-oriented generation alone does not ensure that generated candidates remain editable, regenerable, and compatible with downstream engineering analysis within native CAD/CAE workflows. This limitation motivates closer integration between generative modeling and editable parametric CAD representations.

\subsection{AI-driven parametric CAD model generation}
\label{subsec:parametric_design}
A parametric CAD model stores not only the final shape but also how it is built: sketches whose primitives are linked by geometric constraints, and a history of features that reference the sketches and one another \cite{seff2020sketchgraphs}. Because of these relationships, a dimensional change propagates predictably to the dependent geometry, which makes the model editable and is commonly referred to as design intent \cite{casey2025aligning}. AI-based CAD generation, which uses learned models to produce CAD data automatically, can therefore be distinguished by how much of this parametric information the generated representation retains. Some approaches focus primarily on geometric and topological representations. Boundary representation approaches, such as SolidGen and BrepGen, generate the faces, edges, and vertices of a solid together with their topological relationships \cite{jayaraman2022solidgen,xu2024brepgen}, and constructive solid geometry approaches, such as CSGNet, CAPRI-Net, and D$^2$CSG, infer the primitives and Boolean operations that reconstruct an input shape \cite{sharma2018csgnet,yu2022capri,yu2023d}. These approaches made it possible to generate or reconstruct solid geometry that CAD systems can directly use. However, their outputs do not necessarily retain the feature history and explicit design variables of a history-based parametric model, and therefore may not directly align with the feature- and dimension-level controls used in conventional parametric CAD workflows.

Other studies address the parametric content directly. At the sketch level, SketchGraphs provided 15 million real-world sketches represented as graphs of primitives and designer-imposed constraints \cite{seff2020sketchgraphs}, and a constraint-generation model was further trained with feedback from a constraint solver so that generated sketches update predictably when dimensions change \cite{casey2025aligning}. At the model level, DeepCAD represented a model as a sequence of sketch and extrusion commands and learned a latent space for generating new parametric models \cite{wu2021deepcad}, SkexGen separated topology, geometry, and extrusion into disentangled codebooks to control the generated models \cite{xu2022skexgen}, WHUCAD extended the command vocabulary to features such as revolves and fillets \cite{fan2025parametric}, and GenCAD-3D generated CAD programs from point clouds and meshes using latent-space alignment and diffusion \cite{yu2026gencad}. Text and images have also been used as inputs to generate command sequences or executable CAD code \cite{khan2024text2cad,li2025llm4cad,doris2026cad}. Together, these studies made it possible to generate editable models with explicit constraints, commands, and parameters. However, much of the existing work has focused on generating new CAD models or programs, while the integration of generated parameters with the feature dependencies and inter-part references of existing industrial assemblies has received comparatively less attention.

Industrial design changes, by contrast, are often implemented within native parametric assemblies that have already been established, where feature history, parent--child references, and geometry shared among parts determine how dimensional changes propagate. This study shares the goal of editable parametric output but differs in what is generated: it does not generate a new CAD representation. Instead, it generates design variables mapped to an existing CATIA tailgate assembly so that the original modeling history and reference structure can be retained during design exploration. Such a representation provides a practical means of connecting generative design exploration with subsequent detailed engineering activities in the native CAD/CAE workflow.

\subsection{Surrogate-assisted design optimization}
\label{subsec:surrogate_assisted_design_optimization}
Surrogate-assisted design optimization replaces repeated expensive engineering analyses with computationally efficient predictive models, enabling a larger number of candidate designs to be evaluated during optimization \cite{forrester2008engineering}. GPR, neural networks, and related surrogate models have therefore been widely coupled with sequential and evolutionary optimization algorithms \cite{kim2025surrogate}. In automotive applications, deep learning has also been employed to predict wheel-impact stress responses as an alternative to computationally expensive FEA \cite{shin2023wheel}. In conventional formulations, the surrogate primarily approximates engineering responses while the optimizer determines where candidate designs are evaluated. Bayesian optimization represents a sequential search strategy, whereas evolutionary algorithms such as NSGA-II are widely used for multi-objective design exploration. Surrogate-assisted optimization has also been applied to automotive structural design; for example, GPR and Bayesian optimization have been combined to optimize tailgate rib structures under manufacturing constraints \cite{lee2025bayesian}. Automotive hoods, as representative vehicle closure components, have been studied using learned geometric representations and surrogate models for design exploration and optimization \cite{wollstadt2022carhoods10k}. More recent work has extended this direction to high-fidelity multiphysics datasets and surrogate-based deformation prediction across diverse 3D hood designs \cite{sharma2026autohood3d}.

With the introduction of generative models, the role of the surrogate can extend beyond rapid performance evaluation. Surrogate predictions or gradients can provide performance information to the generative process, allowing generated candidates to be evaluated and directed toward high-performing regions without repeatedly invoking high-fidelity analysis \cite{maze2023diffusion,yang2026physics}. These developments indicate that surrogate models can support design optimization in two distinct but related ways: as efficient performance evaluators for iterative optimization and as performance-information providers within generative design processes. At the same time, surrogate-assisted generative optimization depends not only on surrogate accuracy but also on how effectively performance information is incorporated into generation and training. This issue becomes particularly important when generated candidates move toward sparsely sampled regions of the design space, where local prediction error or surrogate uncertainty may increase \cite{yang2026physics}. In the present study, a common surrogate model supports both performance-oriented generative modeling and comparison with a conventional multi-objective optimizer, followed by native CAD/CAE re-evaluation of selected candidates.

\section{Generative AI-driven parametric design optimization framework}
\label{sec:research_framework}
\begin{figure*}[htb!]
    \centering
        \includegraphics[width=1.00\textwidth]{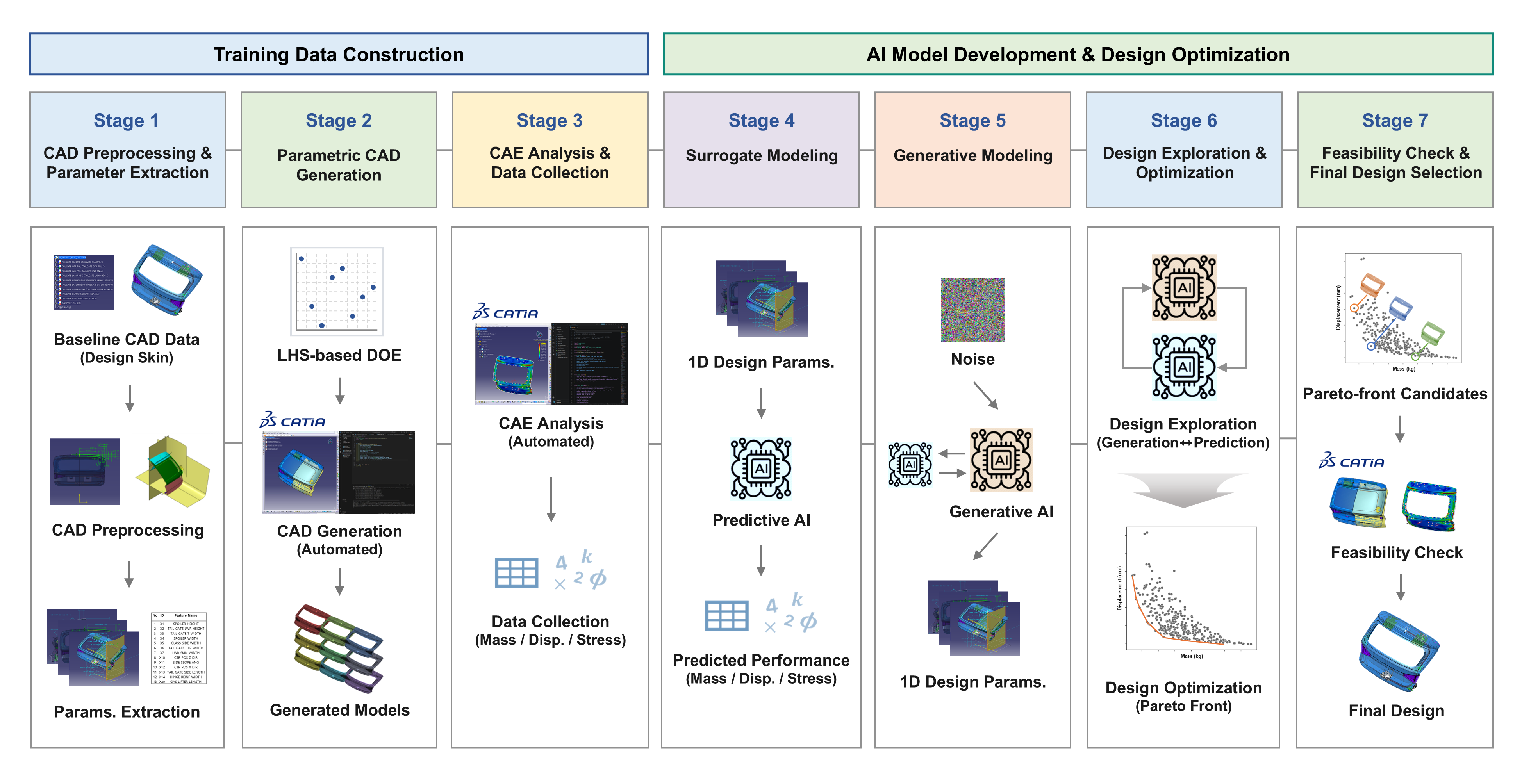}      
    \caption{Overview of the proposed generative AI-driven parametric design optimization framework.}
    \label{fig:research_framework}
\end{figure*}

The proposed generative AI-driven parametric design optimization framework consists of seven stages organized into two sequential parts, as shown in Fig.~\ref{fig:research_framework}: training data construction (Stages~1--3) and AI model development and design optimization (Stages~4--7). The first part establishes an automation-ready parametric design representation and constructs the corresponding CAD/CAE-based engineering dataset. The second part develops predictive and generative AI models, explores the design space toward improved engineering performance, and returns selected candidates to the native CAD/CAE environment for final feasibility assessment.

Stage~1 analyzes the baseline industrial CAD model through CAD preprocessing and parameter extraction, restructuring it into a form suitable for systematic design variation and automation. Design parameters relevant to the engineering problem are identified and extracted to define the numerical design representation used throughout the subsequent stages. Stage~2 generates parametric CAD models by sampling the design space through design of experiments and automatically mapping the sampled parameter combinations back to the parametric CAD model. Stage~3 evaluates the generated CAD models through automated structural CAE analysis and constructs design--performance pairs containing the design parameters and corresponding engineering responses, including mass, maximum displacement, and maximum von Mises stress.

The resulting dataset is then used for AI model development and design optimization. Stage~4 develops surrogate models that learn the relationship between the one-dimensional design parameters and the engineering responses, providing rapid performance predictions for subsequent design exploration. Stage~5 develops generative models that learn the distribution of successfully realized design parameters and generate new parameter combinations from an initial noise distribution. Engineering performance information provided by the surrogate model is incorporated into both model training and generation to support performance-oriented generation. Stage~6 explores and optimizes candidate designs by coupling generation and prediction and evaluating the candidates in the multi-objective design space to identify Pareto-front candidates. Stage~7 verifies feasibility and selects final designs by mapping selected Pareto-front candidates back to the original CATIA model and re-evaluating them through the native CAD/CAE workflow.

\section{Training data construction through CAD/CAE automation (stages 1 to 3)}
\label{sec:data_construction}

\subsection{CAD preprocessing and parameter extraction (stage 1)}
\label{subsec:cad_preprocessing}
The baseline model is an industrial tailgate CAD model whose model tree was constructed for conventional, designer-driven vehicle development rather than for automated design variation. The model is built on a design skin, the exterior styling surface provided as the design input, and the assembly includes outer and inner panels, reinforcements, and interfaces that share reference geometry and cross-sections. To ensure robust model updates under such variation, the model tree is analyzed to identify the inputs of each feature and the parent--child dependencies through which parameter changes propagate. For this purpose, the CATIA surface-modeling operations used in the model are listed together with their types and input elements, showing how editable dimensions are connected to downstream features and which geometric conditions must be preserved when the model is updated.

\begin{figure*}[htb!]
    \centering
        \includegraphics[width=0.95\textwidth]{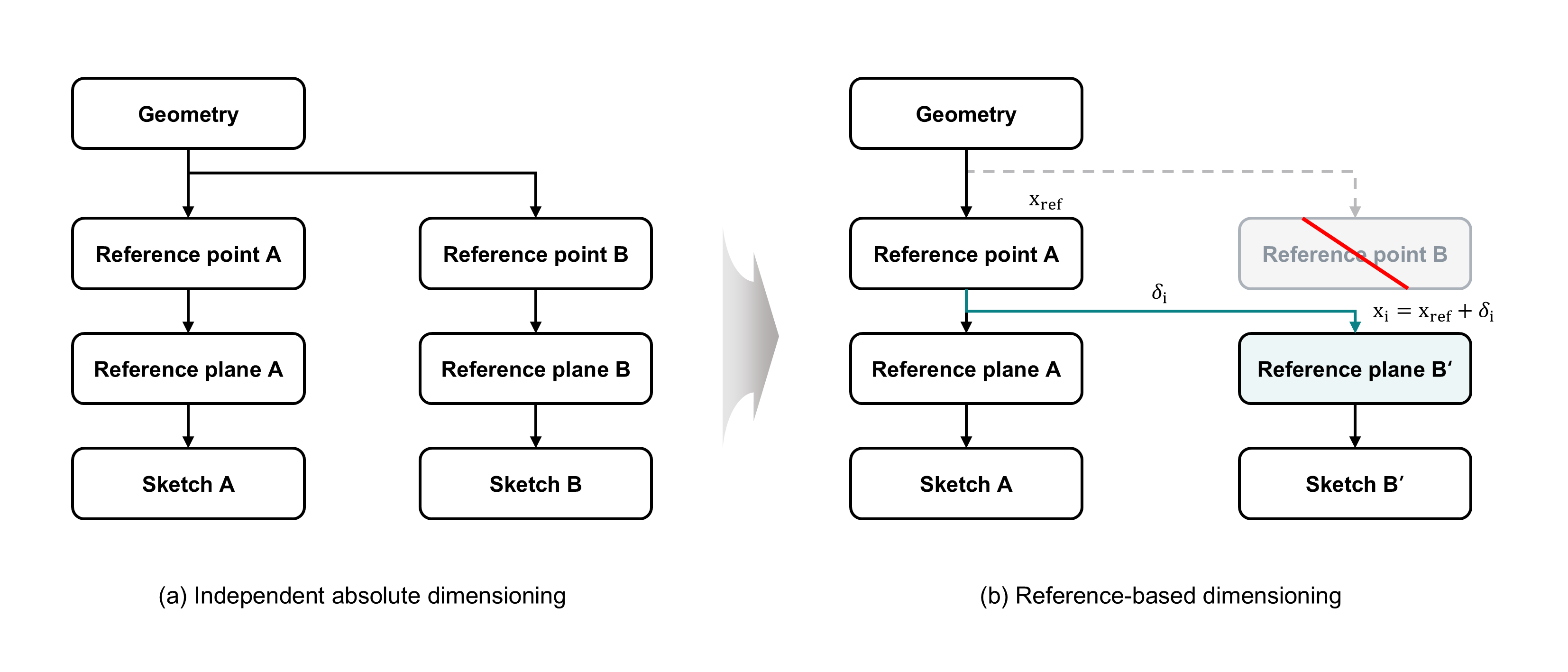}      
    \caption{Conceptual comparison between independent absolute dimensioning and reference-based dimensioning. In the reference-based formulation, editable dimensions are expressed relative to reference locations so that dependent geometry follows changes in upstream features.}
    \label{fig:reference_modeling_concept}
\end{figure*}

Based on these identified dependencies, the model is restructured in collaboration with domain experts according to explicit reference modeling (ERM) principles \cite{bodein2014explicit}. Where appropriate, independently defined absolute dimensions measured from the global origin are replaced by reference-based relations, as illustrated in Fig.~\ref{fig:reference_modeling_concept}, of the form
\begin{equation}
    x_i = x_{\mathrm{ref}} + \delta_i,
    \label{eq:reference_dimension}
\end{equation}
where $x_i$ denotes an editable dimension, $x_{\mathrm{ref}}$ the reference dimension on which it depends, and $\delta_i$ the signed relative offset.

\begin{figure*}[htb!]
    \centering
        \includegraphics[width=0.95\textwidth]{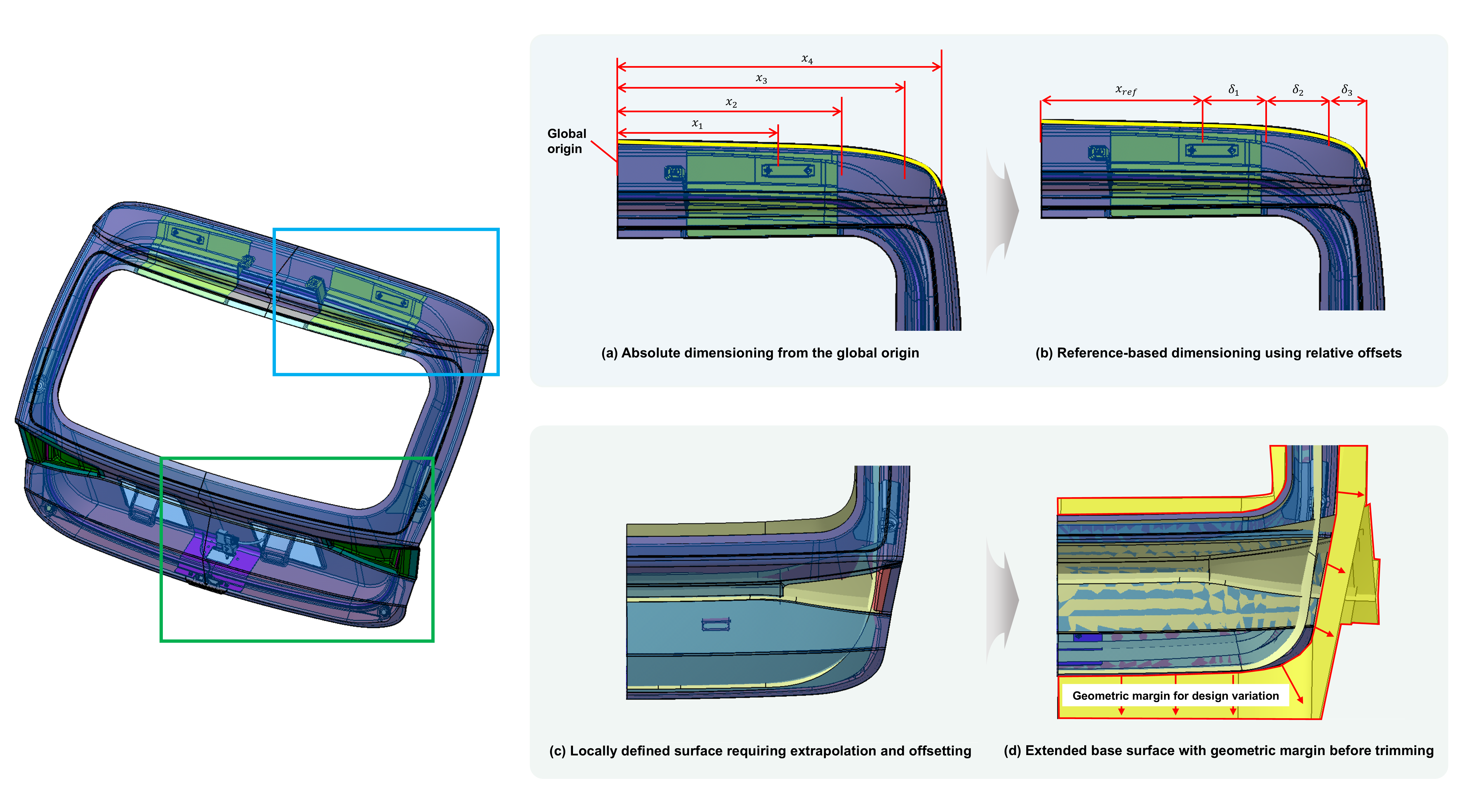}
    \caption{Examples of CAD restructuring for the tailgate case study, including reference-based dimensioning to maintain geometric dependencies during parameter updates and extended base-surface construction to provide sufficient geometric margin for robust design variation.}
    \label{fig:reference_modeling_examples}
\end{figure*}

Dependent geometry thus follows changes in its reference while keeping the specified offset. Surface construction is also revised by preparing sufficiently extended base surfaces before trimming to the required regions. Fig.~\ref{fig:reference_modeling_examples} illustrates how these restructuring principles are applied to the tailgate CAD model used as the case study in this work. In the upper row, dimensions independently defined from the global origin are reformulated as relative offsets from a common reference so that related geometry can update consistently. In the lower row, locally defined surfaces requiring subsequent extrapolation and offsetting are replaced by sufficiently extended base surfaces before trimming, thereby providing geometric margin for the prescribed design variation.

Candidate design variables are screened according to their influence on structural performance, their relevance to design requirements, their suitability for automated updating, and the physical realizability and manufacturability of the resulting geometry. To support this selection, each design requirement is checked against the candidate variables it affects, allowing variables that can be changed independently to be distinguished from those determined by other geometry or mounting conditions or required to remain fixed. The selected design variables, $\mathbf{x}=[x_1,\ldots,x_{n}]^\top$, are assigned lower and upper bounds according to design standards and production feasibility. Each selected variable is mapped to its corresponding CAD parameter and parent feature, and the resulting mapping, together with the restructured model, forms the parametric design template used for automated design generation.

\subsection{Automated parametric CAD generation (stage 2)}
\label{subsec:cad_generation}
\begin{figure*}[htb!]
    \centering
        \includegraphics[width=1.00\textwidth]{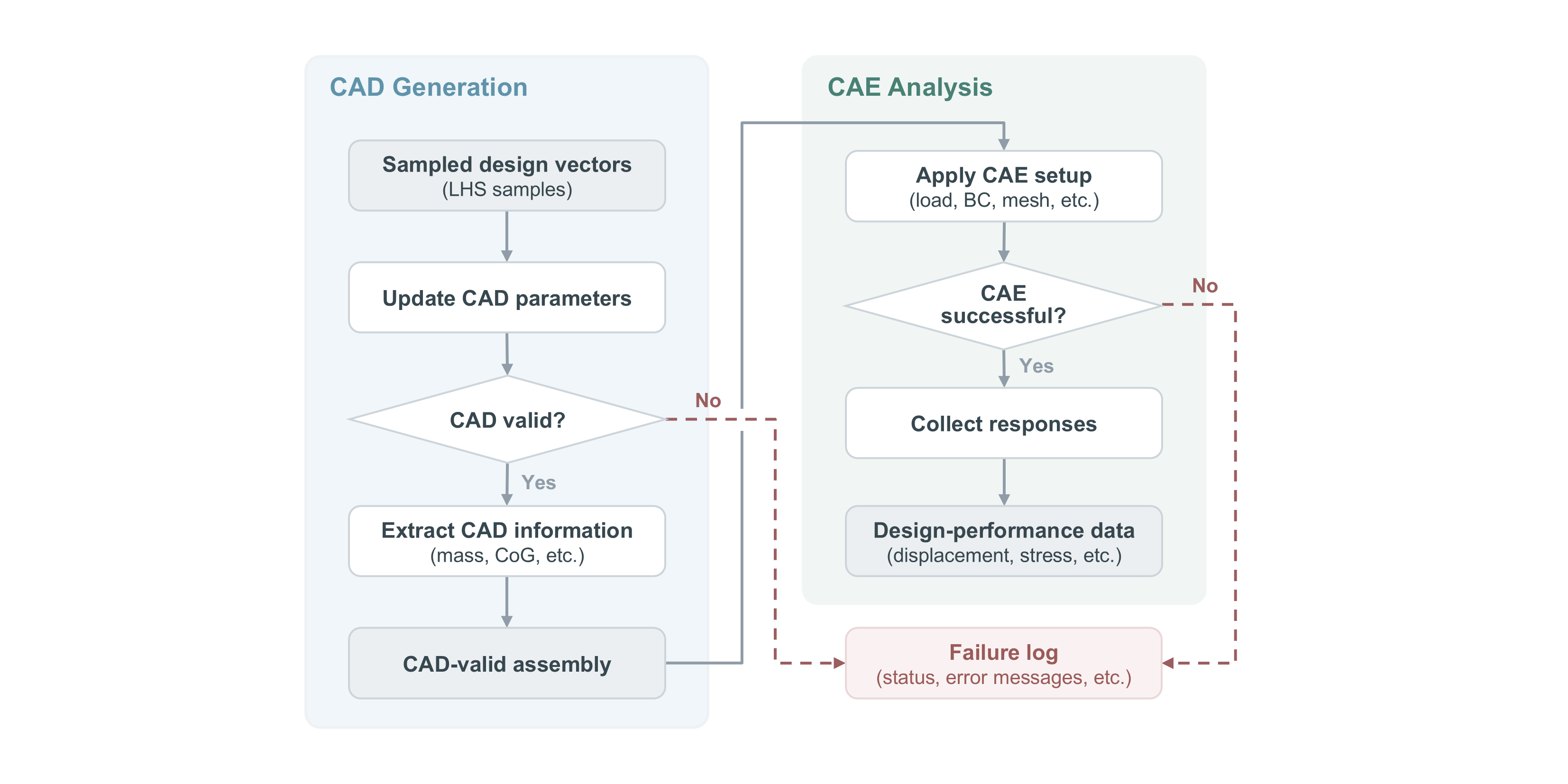}
    \caption{Automated CAD/CAE workflow for constructing the design--performance database, including parametric CAD regeneration, CAE analysis, response extraction, and separate logging of CAD and CAE failures.}
    \label{fig:cad_cae_automation}
\end{figure*}

The design space is sampled using Latin hypercube sampling (LHS) \cite{mckay2000comparison,stein1987large,owen1994controlling}. LHS was selected to provide space-filling coverage of the design space without requiring a full factorial design. Space-filling experimental designs are also advantageous for constructing GPR surrogates, as the distribution and spacing of sample points can affect surrogate-model quality \cite{simpson2001kriging}. For the subsequent GPR modeling, the number of samples is selected with reference to the commonly used rule of thumb of approximately ten runs per input dimension for an initial computer experiment \cite{loeppky2009choosing}. Because satisfying the individual bounds does not guarantee that all dependent features can be constructed, the sampled combinations are treated as candidate inputs whose CAD validity is determined in the subsequent automated generation step. The overall automated CAD/CAE workflow used for data generation is illustrated in Fig.~\ref{fig:cad_cae_automation}.

CAD generation is automated using pyCATIA \cite{pycatia}, a Python interface to the CATIA V5 automation API. The inputs are the baseline assembly document (CATProduct), the parameter-mapping table, and the sampled design vectors. For each candidate, a separate temporary working copy of the baseline model is created so that changes made for one candidate do not carry over to the next. The CATProduct and its referenced part documents (CATPart) are loaded, and an initial assembly update is performed. The mapped CAD parameters are then updated, followed by updates of the affected CATPart documents and the complete CATProduct to determine whether the assembly can be successfully regenerated. For each CAD-valid assembly, the relevant CAD information, including mass, is extracted, the native CAD files are saved for downstream analysis, and the assembly is passed to Stage~3 for structural analysis. For each failed candidate, the parameter combination, generation status, and corresponding error messages are recorded in the failure log, after which processing continues with the next candidate. Each sampled design vector is thus associated with its CAD-generation status and, where available, its native CAD files and extracted information, enabling subsequent analysis of parameter combinations that fail to regenerate.

\subsection{Automated CAE analysis and data collection (stage 3)}
\label{subsec:data_collection}
Successfully regenerated assemblies are evaluated using an automated CAE workflow in the CATIA V5 Generative Assembly Structural Analysis workbench, as summarized in Fig.~\ref{fig:cad_cae_automation}. A torsional loading scenario is defined in collaboration with domain experts, and the same loads, boundary conditions, connections, and mesh settings are applied to all CAD-valid assemblies so that differences in the structural responses can be attributed to geometric variation alone. For each assembly, the analysis document (CATAnalysis) is created and computed, and its completion status is recorded. For successful analyses, the maximum displacement and maximum von Mises stress are collected and stored together with the design variables and the mass obtained during CAD generation. CAD-generation and CAE-analysis failures are recorded separately so that the corresponding failure cases can be examined independently. The cases that successfully complete both CAD generation and CAE analysis provide the design--performance pairs used for subsequent model development.

\section{AI model development and design optimization (stages 4 to 7)}
\label{sec:model_development_design_optimization}

\subsection{Surrogate modeling for engineering performance prediction (stage 4)}
\label{subsec:surrogate_modeling}
GPR has been widely used as a surrogate modeling approach for computationally expensive engineering analyses, building on its probabilistic formulation for computer experiments and its ability to provide flexible nonlinear prediction with uncertainty estimates \cite{sacks1989design,seeger2004gaussian}. GPR, also commonly referred to as Kriging in the context of computer experiments and engineering design, is particularly suitable for problems with a limited number of high-fidelity samples, as encountered in the present study. The selected design variables obtained from the preceding training data construction stages are transformed using min--max normalization before being used as model inputs, and separate GPR models are constructed for the optimization objectives, namely mass and maximum displacement. A Gaussian process defines a distribution over functions as
\begin{equation}
    f(\mathbf{x}) \sim \mathcal{GP}
    \left(m(\mathbf{x}), k(\mathbf{x},\mathbf{x}')\right),
    \label{eq:gpr_prior}
\end{equation}
where $m(\mathbf{x})$ is the mean function and $k(\mathbf{x},\mathbf{x}')$ is the covariance kernel describing the correlation between sampled designs \cite{seeger2004gaussian}. Given the design--performance pairs collected in Stages 1--3, the GPR model provides a predictive distribution for previously unevaluated combinations of the design variables, and the predictive mean is used as the engineering-response estimate in the subsequent design exploration.

The predictive capability of each surrogate is assessed using five-fold cross-validation (CV), with performance evaluated using the coefficient of determination ($R^2$), mean absolute error (MAE), root mean squared error (RMSE), and mean absolute percentage error (MAPE). Several commonly used covariance kernels and associated hyperparameter settings are evaluated, and the configuration with the best overall CV performance is selected for each response. All GPR covariance kernels employ automatic relevance determination (ARD), allowing a separate length scale to be learned for each input dimension. The resulting GPR models are used as computationally efficient performance evaluators during performance-oriented generative modeling and multi-objective design exploration. To further assess their predictive reliability for newly generated designs, surrogate predictions for selected Pareto-region candidates are compared with native CAE results in Stage~7.

\subsection{Performance-oriented generative modeling (stage 5)}
\label{subsec:generative_modeling}
The generative models operate on the same normalized design-variable representation used by the surrogate models. Rather than directly generating CAD geometry, they learn the distribution of design variables obtained through the preceding CAD/CAE data-construction workflow. The generated variables can therefore be mapped back to the feature-based CAD template and reconstructed in the native CAD environment.

Two generative model families, diffusion and flow matching, are considered. Because their mathematical foundations are well established in the literature, only their basic formulations relevant to the present study are described here. Both models are trained using the same normalized design variables to enable a consistent comparison. Diffusion models learn to reverse a progressive noising process so that samples drawn from a Gaussian distribution can be transformed into samples following the target data distribution \cite{ho2020denoising,song2020score}. In the commonly used noise-prediction formulation, the diffusion model is trained by minimizing
\begin{equation}
    L_{\mathrm{DM}} =
    \mathbb{E}\left[
    \left\|
    \boldsymbol{\epsilon}
    -
    \boldsymbol{\epsilon}_{\theta}(\mathbf{x}_{t},t)
    \right\|^{2}
    \right],
    \label{eq:diffusion_loss}
\end{equation}
where $\mathbf{x}_{t}$ denotes the noisy design-variable representation at diffusion step $t$, $\boldsymbol{\epsilon}$ is the added Gaussian noise, and $\boldsymbol{\epsilon}_{\theta}$ is the noise predicted by the diffusion model. Flow matching instead learns a continuous velocity field that transports samples from a simple source distribution to the target data distribution through an ordinary differential equation \cite{lipman2022flow,tong2023improving}. The flow matching model is trained by minimizing
\begin{equation}
    L_{\mathrm{FM}} =
    \mathbb{E}\left[
    \left\|
    \mathbf{v}_{\theta}(\mathbf{x}_{t},t)
    -
    \mathbf{u}_{t}(\mathbf{x}_{t})
    \right\|^{2}
    \right],
    \label{eq:flow_matching_loss}
\end{equation}
where $\mathbf{x}_{t}$ denotes an intermediate design-variable representation along the probability path at continuous time $t$, $\mathbf{v}_{\theta}$ is the learned velocity field, and $\mathbf{u}_{t}$ is the target velocity field defined by the probability path. In the subsequent performance-oriented formulation, $L_{\mathrm{base}}$ denotes the corresponding model-specific loss, i.e., $L_{\mathrm{DM}}$ for the diffusion model and $L_{\mathrm{FM}}$ for the flow matching model. 

\begin{figure*}[htb!]
    \centering
    \includegraphics[width=0.95\textwidth]{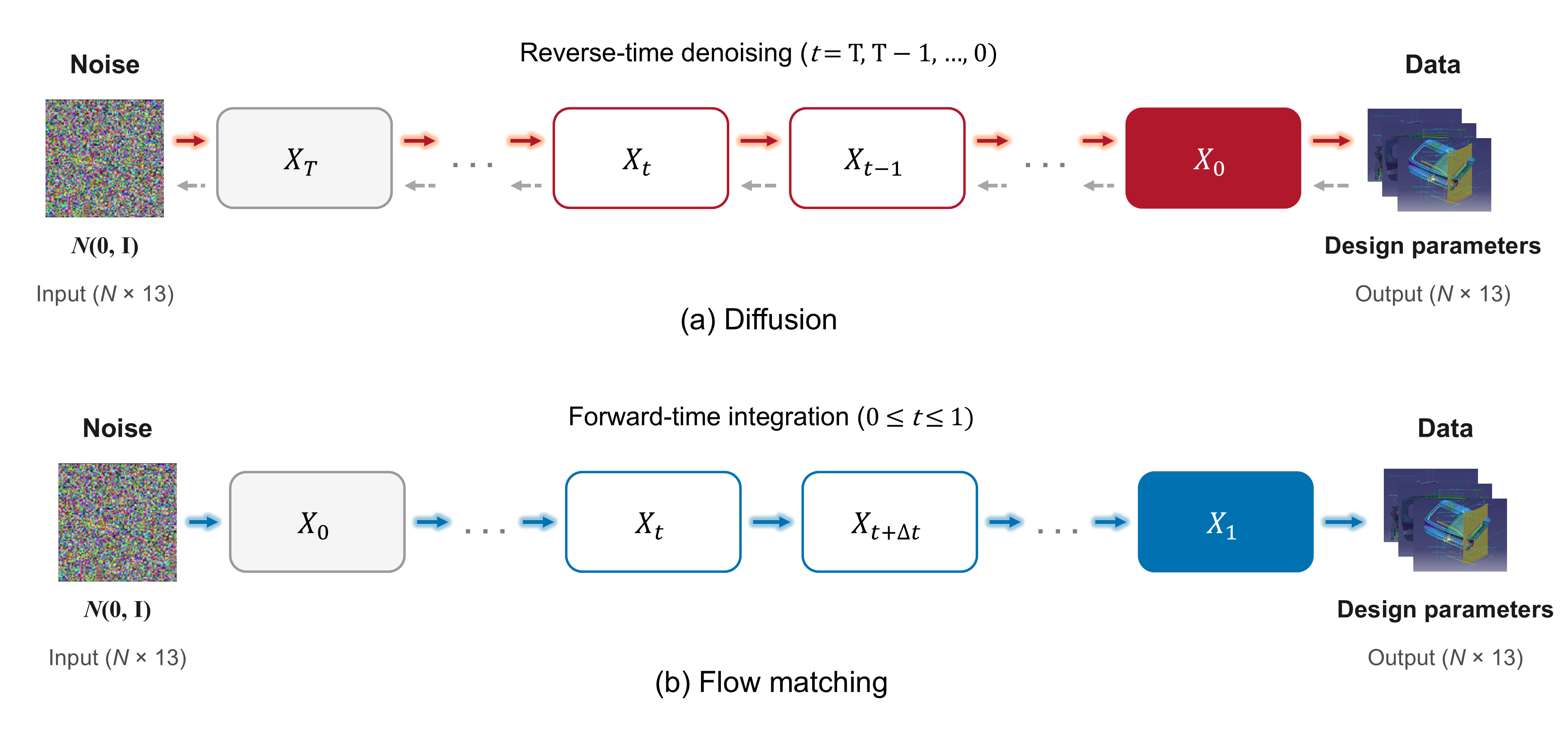}
    \caption{Generation processes of the two generative model families on the normalized design-variable space: (a) reverse-time denoising in the diffusion model and (b) forward-time integration of the learned velocity field in the flow matching model. Both models transform samples from a standard normal distribution into normalized 13-dimensional design-variable vectors.}
    \label{fig:generative_architecture}
\end{figure*}

Fig.~\ref{fig:generative_architecture} summarizes the generation processes of the two generative model families considered in this study. Both models operate on the normalized design-variable representation and are initialized with samples drawn from a standard normal distribution. For the diffusion model, the forward process progressively perturbs data with noise, whereas generation proceeds in the reverse direction from $\mathbf{x}_{T}$ to $\mathbf{x}_{0}$ through iterative denoising. In the noise-prediction formulation adopted here, the trained network predicts the noise component associated with each noisy state and uses this information during the reverse process. In contrast, flow matching learns a time-dependent velocity field that transports samples continuously from a simple source distribution to the target data distribution. Generation is then performed by numerically integrating the learned velocity field from $t=0$ to $t=1$. In both cases, the resulting samples are normalized 13-dimensional design-variable vectors that can be mapped back to the parametric CAD template.

The baseline generative models are trained without explicit engineering-performance information, whereas the performance-oriented variants incorporate information from the GPR surrogate models developed in Stage~4. A common performance objective is defined as
\begin{equation}
    J(\mathbf{x}) =
    w_m \frac{\hat{m}(\mathbf{x})}{s_m}
    +
    w_d \frac{\hat{d}(\mathbf{x})}{s_d},
    \label{eq:performance_objective}
\end{equation}
where $\hat{m}(\mathbf{x})$ and $\hat{d}(\mathbf{x})$ denote the surrogate-predicted mass and maximum displacement, $s_m$ and $s_d$ scale the two responses to comparable magnitudes, and $w_m$ and $w_d$ determine their relative contributions.

Engineering performance is incorporated through two complementary mechanisms. First, performance-guided generation uses the normalized descent direction of the surrogate-based performance objective. The guidance direction and the scaled guidance vector are defined as
\begin{equation}
    \mathbf{g}_{\mathrm{guide}}
    =
    -\frac{\nabla J(\hat{\mathbf{x}})}
    {\left\|\nabla J(\hat{\mathbf{x}})\right\|},
    \label{eq:performance_guidance_direction}
\end{equation}
\begin{equation}
    \mathbf{p}_{\mathrm{guide}}
    =
    \kappa \mathbf{g}_{\mathrm{guide}},
    \label{eq:performance_guidance_scale}
\end{equation}
where $\hat{\mathbf{x}}$ denotes the estimate of the generated design, $\nabla J(\hat{\mathbf{x}})$ is the gradient of the surrogate-based performance objective with respect to the estimated design variables, $\|\cdot\|$ denotes the Euclidean norm, and $\kappa \geq 0$ controls the guidance scale. Thus, $\mathbf{g}_{\mathrm{guide}}$ represents a unit descent direction for decreasing the surrogate-based performance objective, whereas $\mathbf{p}_{\mathrm{guide}}$ represents the scaled guidance used to steer the generative process toward lower predicted objective values. During sampling, $\mathbf{p}_{\mathrm{guide}}$ is incorporated into the model-specific state update at the prescribed guidance intervals. It biases the reverse denoising trajectory of the diffusion model and the forward transport trajectory of the flow matching model toward design regions with lower surrogate-predicted objective values. The approach follows the general principle of gradient-based controllable generation, in which differentiable objectives or external guidance functions are used to direct diffusion- and flow-based generation \cite{chung2022diffusion,bansal2024universal,ben2024d}. Surrogate-based guidance has also been applied in engineering generative design to explicitly direct generated designs toward improved structural performance \cite{maze2023diffusion}.

Second, performance-aware training incorporates the same surrogate-based objective into the training loss. The performance term and total loss are defined as
\begin{equation}
    L_{\mathrm{performance}}
    =
    J(\hat{\mathbf{x}}),
    \label{eq:performance_loss}
\end{equation}
\begin{equation}
    L_{\mathrm{total}}
    =
    L_{\mathrm{base}}
    +
    \lambda L_{\mathrm{performance}},
    \label{eq:total_loss}
\end{equation}
where $\lambda \geq 0$ denotes the loss weight controlling the contribution of the performance term during training. The effects of performance-guided generation and performance-aware training are evaluated individually and in combination, together with the sensitivity to the guidance scale and loss weight.

The principal model and training hyperparameters are tuned before the comparative experiments using a consistent model-selection protocol. The selected configurations are then fixed for the subsequent ablation and optimization comparisons, while detailed implementation settings are reported in the experimental setup.

\subsection{Multi-objective design exploration and optimization (stage 6)}
\label{subsec:design_exploration}
The structural design problem is formulated as the following multi-objective optimization problem:
\begin{equation}
\begin{aligned}
    \text{find} \quad
    & \mathbf{x} = [x_1,x_2,\ldots,x_n] \\[1mm]
    \min_{\mathbf{x}} \quad
    & \mathbf{f}(\mathbf{x})
    = \left[m(\mathbf{x}),d(\mathbf{x})\right] \\[1mm]
    \text{s.t.} \quad
    & \mathbf{x}^{L} \leq \mathbf{x} \leq \mathbf{x}^{U} \\[1mm]
    & \sigma(\mathbf{x}) \leq \sigma_{\mathrm{allow}}
\end{aligned}
\label{eq:multiobjective_problem}
\end{equation}
where $\mathbf{x}$ denotes the design variables, $\mathbf{x}^{L}$ and $\mathbf{x}^{U}$ are their lower and upper bounds, and $m(\mathbf{x})$, $d(\mathbf{x})$, and $\sigma(\mathbf{x})$ denote mass, maximum displacement, and von Mises stress, respectively. The allowable stress $\sigma_{\mathrm{allow}}$ is defined according to the internal engineering design criterion. Absolute design-variable bounds and the allowable stress are not disclosed because of proprietary restrictions, and the optimization results are therefore reported using normalized design variables where necessary. During design exploration, mass and maximum displacement are evaluated using the GPR surrogate models developed in Stage~4. Because no stress surrogate is used during Stage~6, the stress constraint is not imposed during the surrogate-based search; instead, structural stress is evaluated through the native CAE workflow in Stage~7 to determine engineering feasibility. Selected Pareto-region candidates are reconstructed and re-evaluated, and only designs satisfying the allowable stress criterion are retained in the final feasible design set.

NSGA-II is employed as a conventional multi-objective optimization baseline \cite{deb2002fast}. The generative search methods and NSGA-II are compared under matched surrogate-evaluation budgets and repeated independent runs to enable a consistent assessment of their search performance. Pareto quality is evaluated using hypervolume (HV), the number of nondominated solutions, maximum spread, and related performance indicators \cite{zitzler2003performance}. This comparison examines whether learning the distribution of previously realized designs can support efficient search under limited evaluation budgets and how the relative performance changes as additional evaluations become available.

\subsection{CAD/CAE feasibility check and final design selection (stage 7)}
\label{subsec:feasibility_check_design_selection}
Selected Pareto-region candidates are mapped back to the original CATIA parametric template and reconstructed in the native CAD environment. For each candidate, the affected parts and assembly are updated, followed by structural analysis using the CAE workflow. Feasibility is assessed sequentially based on successful CAD regeneration, successful CAE analysis, and satisfaction of the allowable stress criterion, and candidates that fail any of these checks are excluded from the final feasible candidate set. For successfully verified candidates, the surrogate-predicted mass and maximum displacement are compared with the corresponding native CAD/CAE results to assess prediction accuracy for designs obtained through the optimization process. The resulting feasible candidates are then used for final design selection based on their multi-objective performance and engineering feasibility.

\section{Industrial case study: automotive tailgate assembly design}
\label{sec:case_study}

\subsection{Vehicle closure system and tailgate design problem}
\label{subsec:problem_summary}
\begin{figure*}[htb!]
    \centering
        \includegraphics[width=1.00\textwidth]{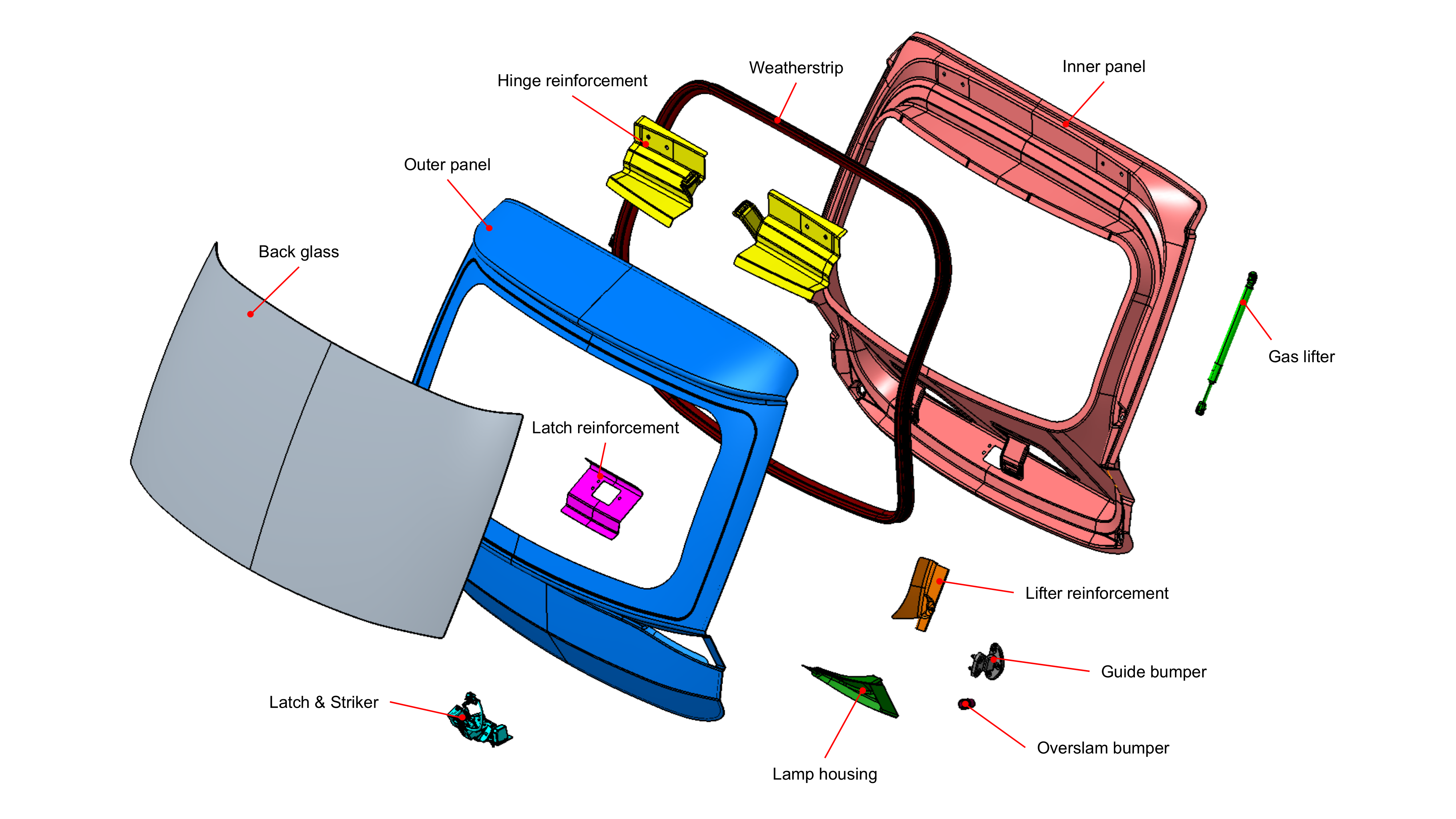}
    \caption{Exploded view of the major structural and functional components of the automotive tailgate assembly used in the case study.}
    \label{fig:tailgate_components}
\end{figure*}

Vehicle closure systems comprise the movable structures and operating mechanisms that open and close vehicle access openings, including doors, hoods, trunk lids, and tailgates \cite{malen2020fundamentals}. They must satisfy requirements on mass, structural stiffness, opening and closing performance, sealing, and noise and vibration \cite{lee2025design}, and these requirements often conflict, as reported for rattle noise and closure effort in tailgates \cite{lee2025multidisciplinary} and for wind noise and squeaking in door weatherstrips \cite{lee2023multidisciplinary}. Among these systems, the present study focuses on the tailgate, whose structural design must provide sufficient stiffness while satisfying weight and packaging requirements. A typical tailgate assembly consists of outer and inner panels, back glass, weatherstrip, hinge, latch, and lifter reinforcements, together with closure hardware and supporting components such as the latch and striker, gas lifter, guide bumper, and overslam bumper, as illustrated in Fig.~\ref{fig:tailgate_components}. The joined panel structure contributes substantially to the overall stiffness, while the reinforcements support local mounting regions and transfer concentrated loads to the surrounding panels \cite{malen2020fundamentals,lee2025multidisciplinary}. Accordingly, the geometry and cross-sectional characteristics of these structural members are important design factors, and an important design objective is to maintain the required structural stiffness while reducing mass.

These structural relationships are also reflected in the feature-based CAD model, where the panels and reinforcements are not defined independently. The design skin and the master references built on it, such as parting lines and section definitions, are shared across multiple components, so that changes to an upstream reference can propagate to downstream features and parts. Consequently, design exploration must consider not only structural performance but also whether the modified assembly can be successfully regenerated and analyzed in the native CAD/CAE environment. The present study therefore employs a research model derived from an industrial tailgate CAD model, with selected geometric details modified to protect vehicle-specific design information. The case study evaluates mass and structural responses under torsional loading.

\subsection{Experimental setup}
\label{subsec:experimental_setup}
\begin{figure*}[htb!]
    \centering
        \includegraphics[width=1.00\textwidth]{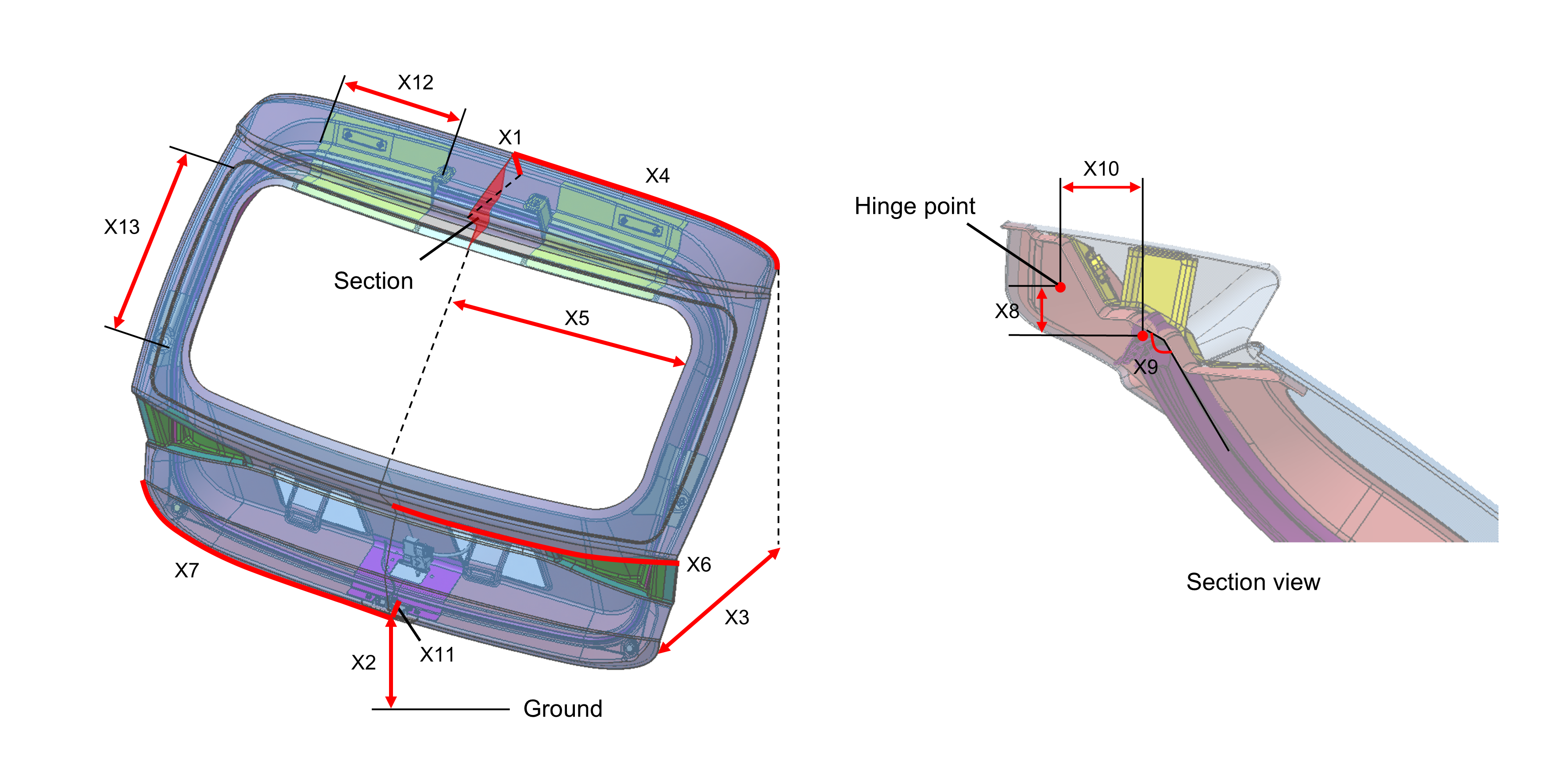}
    \caption{Geometric locations of the 13 design variables selected for the tailgate case study. Absolute dimensional values and design-variable bounds are omitted for proprietary reasons.}
    \label{fig:tailgate_design_variables}
\end{figure*}

The case study uses one industrial tailgate design skin and the corresponding CATIA V5 assembly. The CAD model was preprocessed based on ERM principles to preserve geometric dependencies during automated design variation. Thirteen design variables were selected from approximately 130 raw CAD parameters in collaboration with domain experts, following the screening procedure described in Stage~1. The selected variables represent principal geometric characteristics of the outer and inner panels, back glass, weatherstrip, hinge reinforcement, and gas lifter. Their definitions are summarized in Table~\ref{tab:tailgate_design_variables}, and their geometric locations are illustrated in Fig.~\ref{fig:tailgate_design_variables}. The lower and upper bounds of the variables were determined with domain experts according to design standards and mass-production feasibility. Before surrogate and generative model development, each variable is normalized to $[0,1]$ using its prescribed lower and upper bounds.

\begin{table*}[htb!]
    \centering
    \caption{Definition of the 13 geometric design variables used in the tailgate case study.}
    \label{tab:tailgate_design_variables}
    \begin{tabular}{lcc}
        \toprule
        \textbf{Design variable} & \textbf{Associated component} & \textbf{Geometric parameter} \\
        \midrule
        $x_1$  & Outer panel         & Spoiler-region height \\
        $x_2$  & Outer panel         & Lower-section height \\
        $x_3$  & Outer panel         & Longitudinal length \\
        $x_4$  & Outer panel         & Spoiler-region width \\
        $x_5$  & Back glass          & Width \\
        $x_6$  & Outer panel         & Center-section width \\
        $x_7$  & Outer panel         & Lower-section width \\
        $x_8$  & Weatherstrip        & Reference position along the $Z$-axis \\
        $x_9$  & Inner panel         & Upper-section inclination angle \\
        $x_{10}$ & Weatherstrip      & Reference position along the $X$-axis \\
        $x_{11}$ & Inner panel       & Lower-side length \\
        $x_{12}$ & Hinge reinforcement & Width \\
        $x_{13}$ & Gas lifter        & Length \\
        \bottomrule
    \end{tabular}
\end{table*}

The initial design of experiments consists of 200 LHS combinations, corresponding to approximately 15 samples per design variable and following the sample-size guideline adopted in Stage~2. Of these, 189 cases successfully complete both CAD generation and CAE analysis and provide valid design--performance pairs. Only data obtained successfully through the predefined automated CAD/CAE workflow are retained for subsequent model development; failed cases are not manually repaired or supplemented. This maintains consistent data provenance, avoids operator-dependent intervention, and allows the resulting dataset to reflect the practical feasibility and robustness of the automated workflow. The same normalized design-variable representation is used for the subsequent surrogate and generative-model experiments.

Structural evaluation is performed under a torsional loading condition. The upper central region of the tailgate is fully constrained, and asymmetric loads are applied to the lower region. Connections between the outer and inner panels and between the inner panel and reinforcements are represented using spot-weld and seam-weld connections. The nominal mesh sizes are 4~mm for the outer panel, 12~mm for the inner panel, and 10--12~mm for the reinforcement components, while the mesh sag for the inner panel is set to 1.2~mm. These analysis settings are applied consistently to all sampled designs. Mass and maximum displacement are used as the optimization objectives, while maximum von Mises stress is used as an engineering feasibility criterion and evaluated against an internal design criterion specified by the company during native CAE verification.

Because the case study is based on an industrial product-development model, detailed vehicle-specific information is disclosed only to the extent permitted by proprietary restrictions. Numerical analysis settings that do not reveal product-specific design information are reported where possible, whereas absolute design-variable values and bounds, vehicle-specific load magnitudes, detailed material specifications, and internal design criteria are not disclosed. These quantities are held fixed throughout the comparative analyses, and normalized variables or relative performance measures are reported where appropriate.

The principal model architectures and training settings used in the case study are summarized in Table~\ref{tab:model_configuration}. The surrogate configurations were selected based on repeated cross-validation, while the generative-model architectures and training hyperparameters were selected before the comparative experiments using validation performance. The selected configurations were subsequently fixed across the main comparisons, except for the performance-incorporation parameters explicitly varied in the ablation and sensitivity analyses. Both generative model families were evaluated using the same data split and seed set.

\begin{table*}[htb!]
    \centering
    \caption{Principal model architectures and training settings used in the case study.}
    \label{tab:model_configuration}

    \small
    \setlength{\tabcolsep}{5pt}
    \renewcommand{\arraystretch}{1.12}

    \begin{tabularx}{\textwidth}{
        @{}
        >{\raggedright\arraybackslash}p{0.25\textwidth}
        >{\raggedright\arraybackslash}X
        >{\raggedright\arraybackslash}X
        @{}
    }
        \toprule
        \textbf{Item}
        & \multicolumn{2}{c}{\textbf{Adopted configuration}} \\
        \midrule

        \multicolumn{3}{l}{\textbf{GPR surrogate models (Stage~4)}} \\
        \addlinespace[2pt]

        Input
        & \multicolumn{2}{l}{13 normalized design variables} \\

        Response
        & Mass
        & Maximum displacement \\

        Training / test split
        & \multicolumn{2}{l}{151 training / 38 test designs} \\

        Covariance kernel
        & Mat\'ern ($\nu=5/2$) + rational quadratic (ARD)
        & Mat\'ern ($\nu=1/2$) (ARD) \\

        Model selection
        & \multicolumn{2}{l}{Repeated five-fold cross-validation} \\

        \midrule

        \multicolumn{3}{l}{\textbf{Generative models (Stage~5)}} \\
        \addlinespace[2pt]

        Model
        & Diffusion
        & Flow matching \\

        Input
        & \multicolumn{2}{l}{13 normalized design variables} \\

        Backbone
        & \multicolumn{2}{l}{MLP with sinusoidal time embedding, layer normalization, and SiLU activation} \\

        Hidden width / layers
        & 512 / 3
        & 128 / 3 \\

        Time-embedding dimension
        & 256
        & 192 \\

        Training formulation
        & 1000 forward-noising steps with a linear $\beta$ schedule
        & Conditional flow matching ($\sigma = 0$) \\

        Optimizer
        & \multicolumn{2}{l}{AdamW} \\

        Learning rate
        & $1.4\times10^{-4}$
        & $8.2\times10^{-4}$ \\

        Batch size / epochs
        & \multicolumn{2}{l}{16 / 1000} \\

        Model selection
        & \multicolumn{2}{l}{Checkpoint with the minimum validation loss} \\

        Generation
        & 100 DDIM denoising steps
        & 100 Euler integration steps \\

        Nominal performance setting
        & \multicolumn{2}{l}{$\lambda=0.05$, $\kappa=0.5$, and $w_m=w_d=1$} \\

        \bottomrule
    \end{tabularx}
\end{table*}

\subsection{Automated CAD/CAE data generation and failure analysis}
\label{subsec:data_generation}
The automated workflow generated usable design--performance pairs for 189 of the 200 attempted LHS points, corresponding to an overall data-acquisition success rate of 94.5\%. The unsuccessful cases provide useful engineering information because they reveal regions where nominal parameter bounds are incompatible with the nonlinear geometric and analysis requirements of the native CAD/CAE workflow. Five of the 200 attempted cases failed during CAD generation. The observed CAD failures occurred in fillet-related operations (2 cases; Fillet and EdgeFillet), followed by Thicksurface (1 case), Corner (1 case), and Point (1 case), in which the intersection point required by a downstream feature was not generated. Analysis of the failed models indicated that rapid curvature transitions can cause offset-surface self-intersection or remove the geometric margin required by downstream operations.

After the 195 CAD-valid cases were passed to structural analysis, six additional cases failed during CAE preparation or meshing, all reporting a topology restoration error during analysis computation. Analysis of these failures indicated that insufficient mesh resolution around highly curved regions was a major contributing factor. Reducing the mesh size and sag improved mesh generation in diagnostic tests; however, these adjusted reruns were used only to investigate the failure mechanism and were not used to supplement the dataset. Only design--performance pairs obtained successfully through the predefined automated CAD/CAE workflow were retained. This maintains consistent data provenance across all samples, avoids operator-dependent intervention, and allows the resulting dataset to reflect the practical robustness and limitations of the automated workflow itself. These observations also illustrate why a nominal box-bounded DOE should not be assumed to define an entirely valid industrial design space. The outcome of each workflow step is summarized in Table~\ref{tab:data_generation_outcome}.

\begin{table*}[htb!]
    \centering
    \caption{Summary of automated CAD/CAE data generation outcomes for the 200 LHS designs.}
    \label{tab:data_generation_outcome}
    \begin{tabular}{lcccc}
        \toprule
        \textbf{Workflow step} & \textbf{Cases processed} & \textbf{Successful cases} & \textbf{Failed cases} & \textbf{Success rate} \\
        \midrule
        CAD generation & 200 & 195 & 5 & 97.5\% \\
        CAE analysis & 195 & 189 & 6 & 96.9\% \\
        \midrule
        Overall CAD/CAE workflow & 200 & 189 & 11 & 94.5\% \\
        \bottomrule
    \end{tabular}
\end{table*}

The failure analysis also influences the interpretation of the AI models. The 189 valid cases represent the successfully realized portion of the sampled design space, so the generative models learn primarily from parameter combinations that have successfully passed through the native CAD/CAE workflow. This provides an implicit bias toward CAD/CAE-realizable regions but does not guarantee that every newly generated combination will remain valid. The direct reconstruction step in Stage~7 is therefore retained even when generated samples lie within all numerical parameter bounds.

\subsection{Engineering response characterization}
\label{subsec:data_analysis}
\begin{figure*}[htb!]
    \centering
    \includegraphics[width=1.00\textwidth]{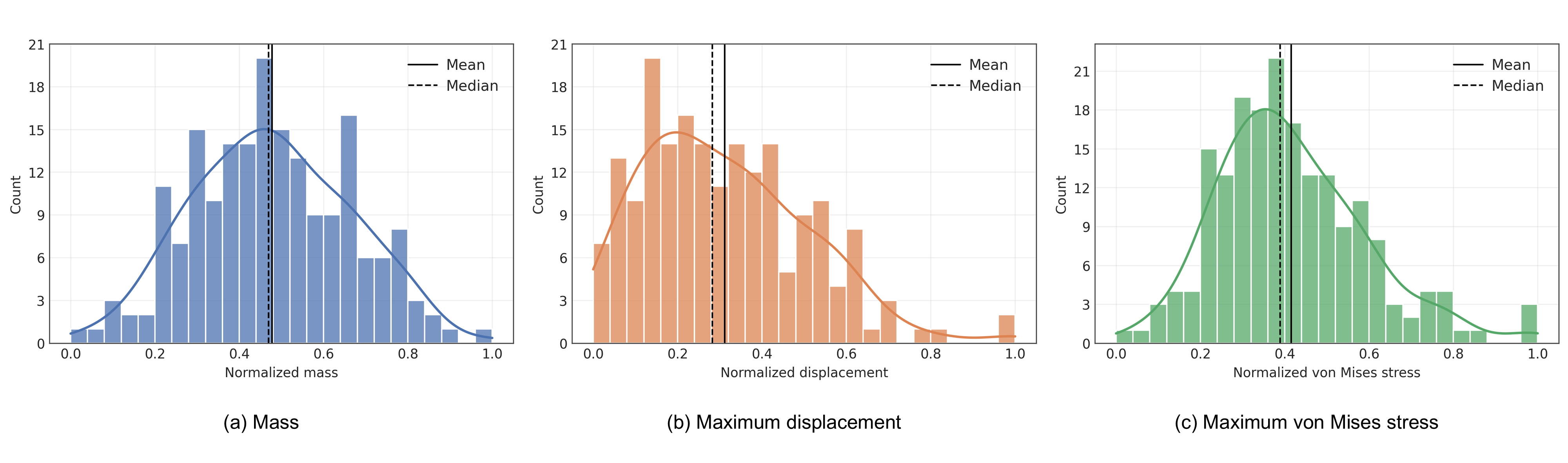}
    \caption{Distributions of the normalized engineering responses obtained from the 189 retained CAD/CAE designs: (a) mass, (b) maximum displacement, and (c) maximum von Mises stress. The solid and dashed vertical lines indicate the mean and median, respectively.}
    \label{fig:data_characterization}
\end{figure*}

Before surrogate and generative model development, the engineering responses of the 189 retained design--performance pairs are examined to characterize the dataset used for subsequent model development (Fig.~\ref{fig:data_characterization}). For proprietary reasons, the response values are normalized to their respective ranges for visualization. The mass response exhibits an approximately symmetric distribution, with similar mean and median values. In contrast, maximum displacement and maximum von Mises stress show more asymmetric distributions with longer upper tails, indicating that a smaller number of designs exhibit substantially larger structural responses. All 189 cases that successfully completed the predefined CAD/CAE workflow are retained for model development without removing cases solely on the basis of statistical extremeness.

The distinct response distributions also motivate separate assessment of the surrogate models for each optimization objective. In particular, maximum displacement exhibits a more pronounced asymmetric distribution and longer upper tail than mass, while their respective predictive characteristics are further examined through the surrogate-model validation in Section~\ref{subsec:surrogate_results}.

\subsection{Surrogate model performance and cross-validation}
\label{subsec:surrogate_results}
Separate GPR surrogate models were developed for the two optimization objectives, mass and maximum displacement. The 189 valid design--performance pairs were randomly divided, with 80\% (151 designs) used for training and 20\% (38 designs) reserved as a hold-out test set. Several covariance kernels, including the radial basis function (RBF), Mat\'ern, rational quadratic, and composite kernels, together with associated hyperparameter configurations, were evaluated using the training set. For each response, the configuration providing the best predictive performance based on repeated five-fold CV was selected as the final surrogate model. The hold-out test set was used only for final performance evaluation. Both inputs and outputs were min--max normalized using the training data to avoid information leakage, and MAE and RMSE were transformed back to the original physical units for reporting. The repeated five-fold CV results are reported as mean $\pm$ standard deviation, while the remaining performance metrics are evaluated on the hold-out test set. The 95\% predictive interval (PI) coverage denotes the fraction of test designs whose CAE values fall within the 95\% PI of the surrogate.

\begin{figure*}[htb!]
    \centering
    \includegraphics[width=0.80\textwidth]{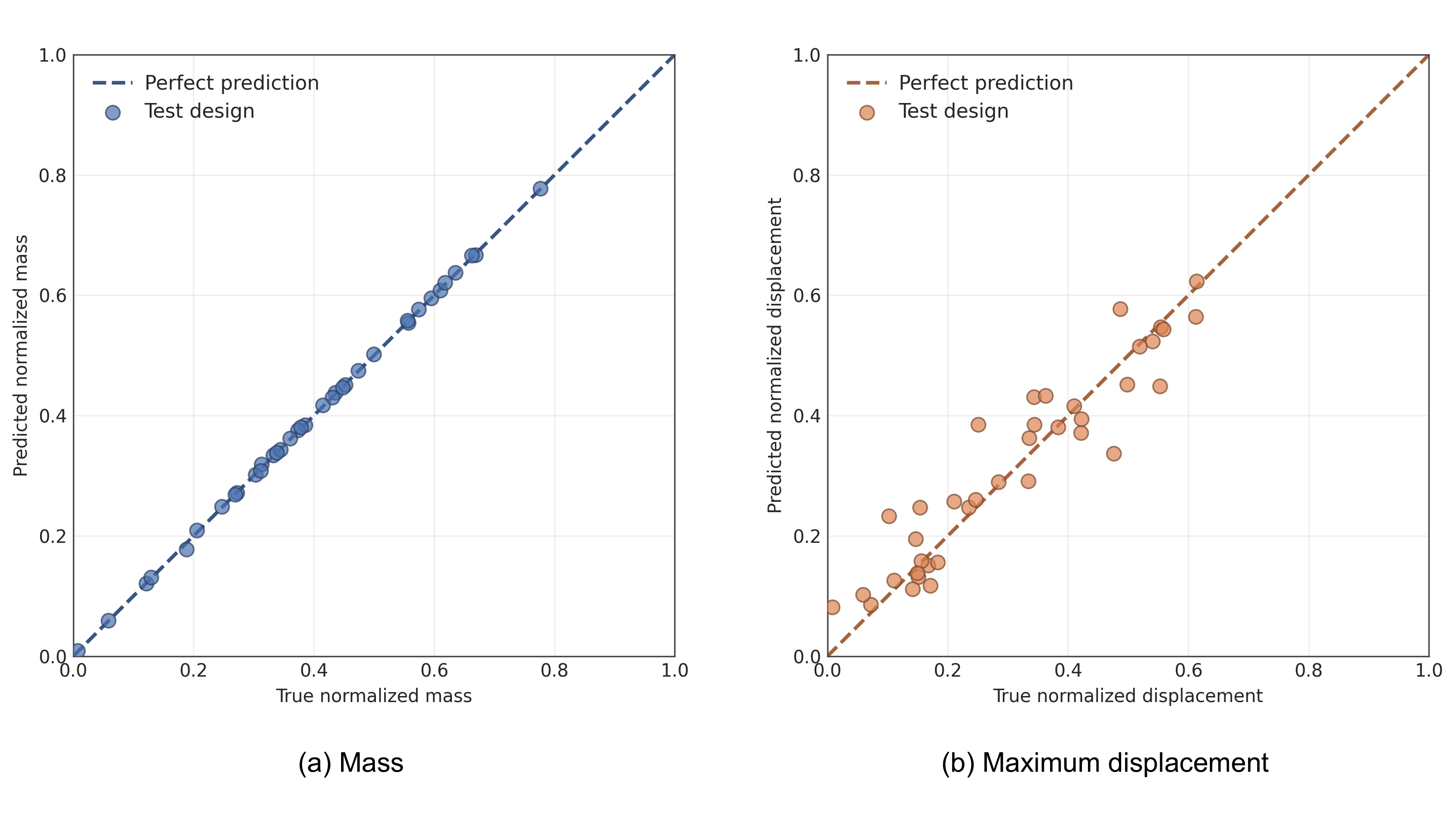}
    \caption{Hold-out parity plots of the GPR surrogate models for (a) mass and (b) maximum displacement using 38 test designs. The axes are normalized using the min--max ranges of the training data.}
    \label{fig:surrogate_validation}
\end{figure*}

Table~\ref{tab:surrogate_performance} and Fig.~\ref{fig:surrogate_validation} show that the two engineering responses exhibit different levels of predictive difficulty. Mass is predicted with very high accuracy, with a hold-out $R^2$ of 0.9998 and a MAPE of 0.03\%. Maximum displacement is more difficult to predict, but the GPR model still achieves a hold-out $R^2$ of 0.8889 and a MAPE of 5.81\%. The repeated CV results show the same overall tendency, indicating that the difference between the two responses is not specific to a particular train--test split. The empirical 95\% PI coverage is 97.4\% for mass and 100.0\% for maximum displacement. These coverage values indicate that the predictive intervals adequately capture the uncertainty of the unseen test designs; they should be interpreted as measures of interval calibration rather than point-prediction accuracy.

\begin{table*}[htb!]
    \centering
    \caption{Predictive performance of the GPR surrogate models for mass and maximum displacement.}
    \label{tab:surrogate_performance}
    \begin{tabular}{lcccccc}
        \toprule
        \textbf{Response} & \textbf{Test $R^2$} & \textbf{5-fold CV $R^2$} & \textbf{MAE} & \textbf{RMSE} & \textbf{MAPE} & \textbf{95\% PI coverage} \\
        \midrule
        Mass & 0.9998 & 0.9996 $\pm$ 0.0001 & 0.0051 & 0.0069 & 0.03\% & 97.4\% \\
        Displacement & 0.8889 & 0.8505 $\pm$ 0.0452 & 0.2564 & 0.3443 & 5.81\% & 100.0\% \\
        \bottomrule
    \end{tabular}
\end{table*}

The difference in predictive accuracy between the two responses is further considered during native CAD/CAE verification. Because maximum displacement exhibits a larger generalization error than mass, the verified candidates are evaluated across the generated Pareto region, with particular attention to the agreement between surrogate predictions and direct CAE results.

\subsection{Performance-oriented generative modeling: ablation and sensitivity}
\label{subsec:generative_results}
For each generative family, the architecture and optimizer hyperparameters of the base model ($\lambda = 0$) were selected using a consistent hyperparameter-search procedure based on validation performance. The selected configurations were then held fixed, and models were trained with the performance-aware loss weight $\lambda \in \{0, 0.01, 0.05, 0.10, 0.20\}$ using five random seeds for each setting. The performance-guidance scale was evaluated at $\kappa \in \{0, 0.25, 0.5, 1.0, 2.0\}$ during the sensitivity analysis. Guided sampling used 100 denoising or integration steps, with guidance applied every second step, and the nominal setting of $\lambda = 0.05$ and $\kappa = 0.5$ was adopted for the main comparisons. The surrogate-evaluation budget was fixed at $N = 500$ for the subsequent analyses, where $N$ denotes the number of candidate designs evaluated by the surrogate model in each run. The same evaluation budget was applied to the generative methods and NSGA-II.

\begin{figure*}[htb!]
    \centering
    \includegraphics[width=1.00\textwidth]{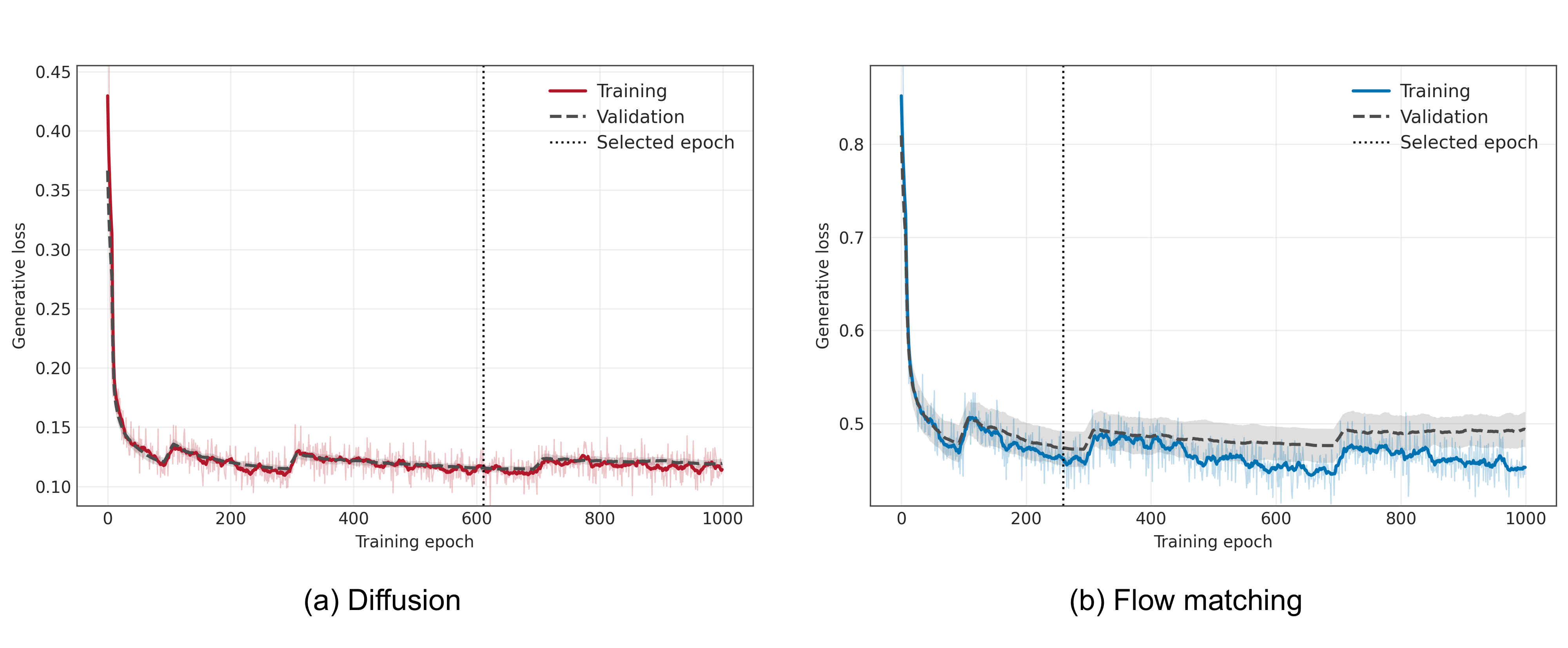}
    \caption{Training and validation losses of the adopted generative-model configurations at $\lambda = 0.05$: (a) diffusion and (b) flow matching. Solid and dashed curves denote the mean training and validation losses over five seeds, respectively; faint curves show the unsmoothed training losses, shaded bands indicate one standard deviation of the validation loss across seeds, and dotted vertical lines mark the mean epoch selected based on the minimum validation loss.}
    \label{fig:learning_curves}
\end{figure*}

Fig.~\ref{fig:learning_curves} shows the training behavior of the adopted generative-model configurations at the nominal performance-aware loss weight of $\lambda = 0.05$. For both model families, the losses decreased rapidly during the early training stage and subsequently stabilized within the 1000-epoch budget. The periodic increases in the loss reflect the warm restarts of the cosine learning-rate schedule. The diffusion model showed close agreement between the training and validation losses, whereas flow matching exhibited a modest train--validation gap during later training. To avoid relying on the final training epoch, the checkpoint with the minimum validation loss was retained for each seed. The selected epoch occurred at 611 on average for diffusion and at 259 for flow matching across the five seeds.

Pareto performance is assessed using three complementary indicators: hypervolume (HV), the number of nondominated (ND) designs, and maximum spread. HV represents the overall quality of the obtained Pareto front by considering the dominated objective space volume (quality), the ND count represents the number of available nondominated design alternatives (quantity), and maximum spread describes the extent of the nondominated solutions in the objective space (diversity). Larger values indicate better performance for all three indicators. HV is computed in the normalized objective space using a common reference point for all methods and runs.

\paragraph{Ablation of the two performance-incorporation mechanisms}
Table~\ref{tab:ablation} summarizes the four variants---Baseline, Performance-guided generation only, Performance-aware training only, and Combined---for diffusion and flow matching. Overall, both performance-incorporation mechanisms improve the Pareto search compared with the baseline generation, although their contributions are different. For diffusion, the combined strategy increases the mean HV from 1.188 to 1.401, corresponding to an improvement of approximately 18\%, while the mean number of nondominated designs increases from 6.4 to 13.4. Performance-guided generation accounts for most of this improvement, whereas performance-aware training provides a smaller stand-alone gain but further improves the consistency of the generated Pareto front when combined with guidance. The improvement was consistently observed across the five random seeds. A similar qualitative trend is observed for flow matching, for which the mean HV increases from 1.159 for the baseline generation to 1.260 for the combined strategy. The magnitude of the improvement is smaller than that observed for diffusion, but the consistent behavior across the two model families indicates that performance-guided generation and performance-aware training provide complementary mechanisms rather than an effect specific to a single generative architecture. Maximum spread does not increase monotonically with Pareto quality, because a wider objective-space extent does not necessarily imply better convergence toward the desirable region. The combined strategy is therefore selected based primarily on its overall Pareto quality and number of nondominated alternatives, while maximum spread is treated as a complementary indicator of diversity rather than the primary selection criterion.

\begin{table*}[htb!]
    \centering
    \caption{Ablation results for performance incorporation at a surrogate-evaluation budget of $N = 500$.}
    \label{tab:ablation}
    \begin{tabular}{llccc}
        \toprule
        \textbf{Family} & \textbf{Method} & \textbf{HV} $\uparrow$ & \textbf{ND count} $\uparrow$ & \textbf{Maximum spread} $\uparrow$ \\
        \midrule
        \multirow{4}{*}{Diffusion}
        & Baseline                               & 1.188 $\pm$ 0.110 & 6.4 $\pm$ 1.8  & 0.424 $\pm$ 0.220 \\
        & Performance-guided generation only     & 1.384 $\pm$ 0.035 & 12.8 $\pm$ 1.6 & 0.591 $\pm$ 0.114 \\
        & Performance-aware training only        & 1.231 $\pm$ 0.088 & 8.4 $\pm$ 1.7  & 0.594 $\pm$ 0.123 \\
        & Combined                               & \textbf{1.401 $\pm$ 0.031} & \textbf{13.4 $\pm$ 2.2} & 0.553 $\pm$ 0.094 \\
        \midrule
        \multirow{4}{*}{Flow matching}
        & Baseline                               & 1.159 $\pm$ 0.046 & 6.6 $\pm$ 2.1  & 0.541 $\pm$ 0.175 \\
        & Performance-guided generation only     & 1.233 $\pm$ 0.039 & 8.0 $\pm$ 2.6  & 0.552 $\pm$ 0.088 \\
        & Performance-aware training only        & 1.195 $\pm$ 0.039 & 7.8 $\pm$ 1.6  & 0.549 $\pm$ 0.082 \\
        & Combined                               & \textbf{1.260 $\pm$ 0.037} & \textbf{10.4 $\pm$ 3.5} & 0.501 $\pm$ 0.075 \\
        \midrule
        \multirow{2}{*}{Reference}
        & NSGA-II                                & 1.378 $\pm$ 0.043 & 34.4 $\pm$ 5.7 & 0.558 $\pm$ 0.105 \\
        & LHS random search                      & 1.243 $\pm$ 0.047 & 7.2 $\pm$ 2.8  & 0.598 $\pm$ 0.023 \\
        \bottomrule
    \end{tabular}
\end{table*}

\begin{figure*}[htb!]
    \centering
    \includegraphics[width=0.95\textwidth]{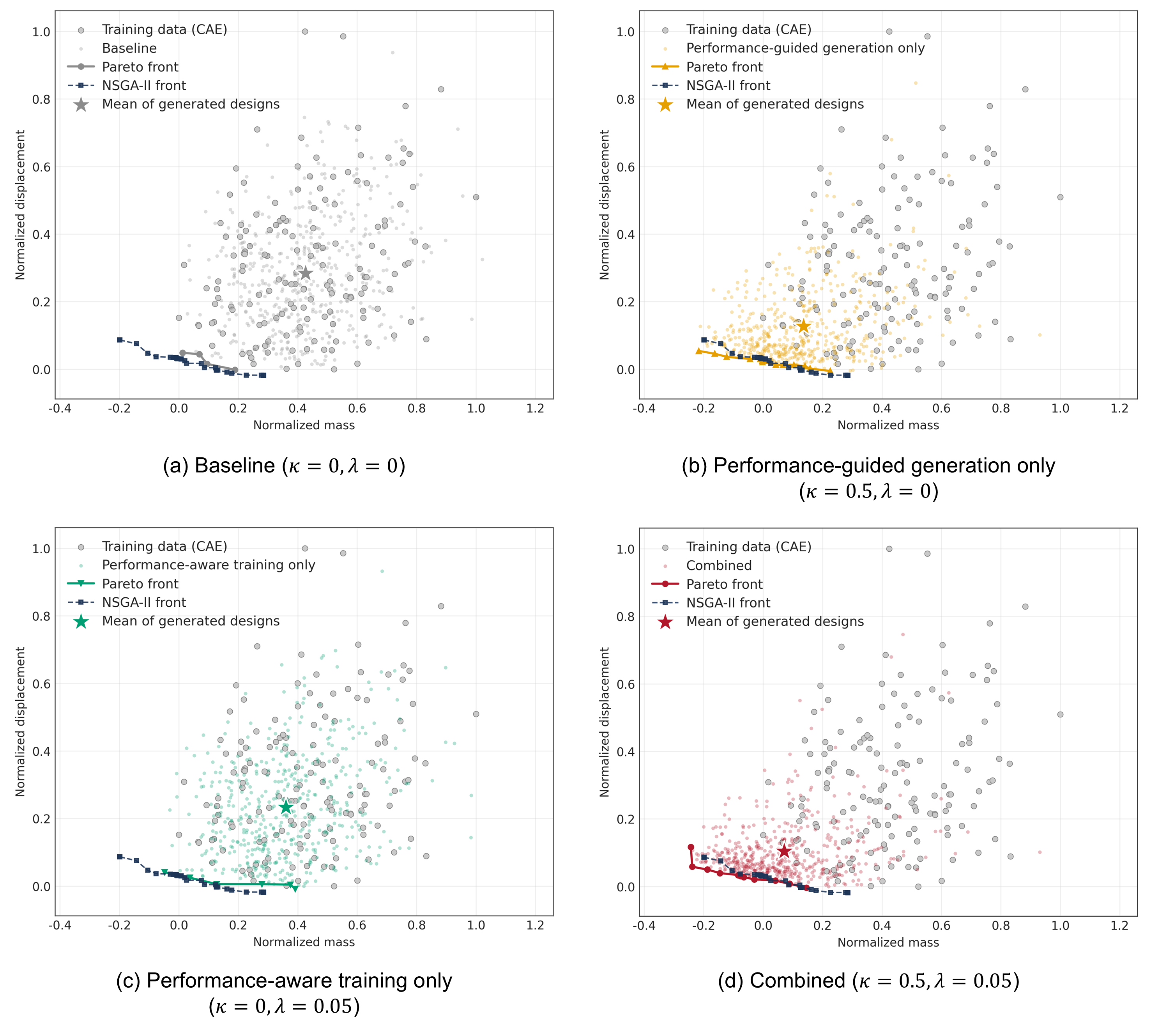}
    \caption{Effect of performance incorporation on the Pareto distributions generated by the diffusion model at $N = 500$: (a) baseline, (b) performance-guided generation only, (c) performance-aware training only, and (d) combined. The generated designs and their nondominated fronts are shown together with the training data, with the NSGA-II front under the same evaluation budget included for reference.}
    \label{fig:ablation_pareto}
\end{figure*}

For visual clarity, Fig.~\ref{fig:ablation_pareto} illustrates the changes in the generated Pareto distribution using diffusion as a representative model. Diffusion is used here as the representative visualization because it shows the clearest response to performance incorporation in Table~\ref{tab:ablation}, while the same qualitative tendency is also observed for flow matching. Baseline generation largely reproduces the region occupied by the training data, with its nondominated front remaining close to the best observed designs. Performance-guided generation shifts the generated candidates toward the low-mass and low-displacement region during sampling, whereas performance-aware training shifts the distribution of generated samples toward the same favorable region even without sampling-time guidance. This indicates that performance information is incorporated into the learned design distribution during training. Their combination produces the most favorable overall Pareto distribution, suggesting that performance-aware training provides a more favorable starting distribution from which surrogate guidance can further direct the generated candidates.

\begin{figure*}[htb!]
    \centering
    \includegraphics[width=0.85\textwidth]{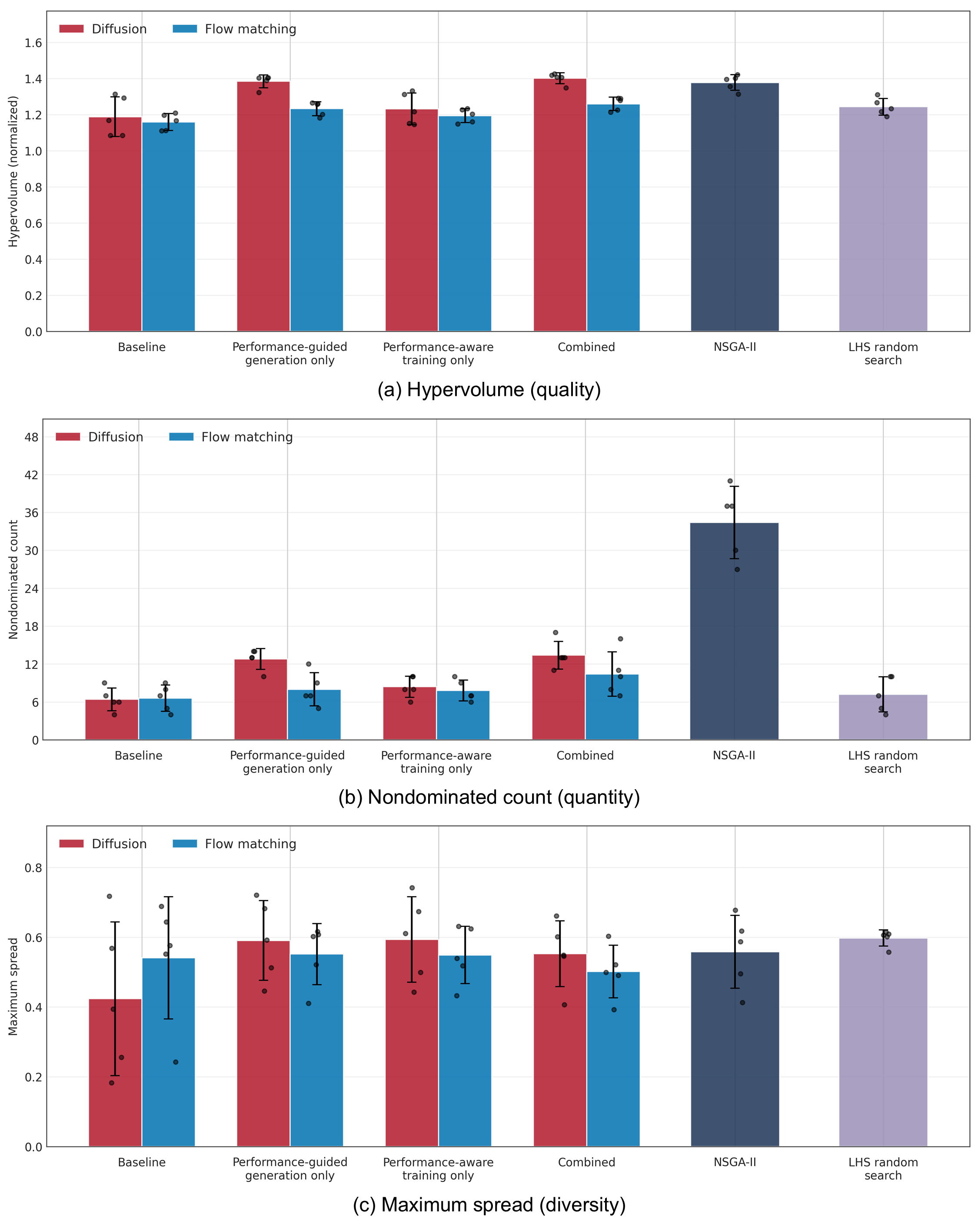}
    \caption{Quantitative ablation results for diffusion and flow matching at $N = 500$: (a) hypervolume, (b) nondominated count, and (c) maximum spread.}
    \label{fig:ablation_metrics}
\end{figure*}

The quantitative comparison in Fig.~\ref{fig:ablation_metrics} shows that the trends observed in the representative Pareto distributions remain consistent across repeated runs and across both generative families. The combined strategy provides the highest mean HV and ND count among the evaluated variants for both diffusion and flow matching, while performance-guided generation contributes most of the improvement relative to the baseline generation. Diffusion shows a stronger response to performance incorporation under the present implementation, whereas flow matching exhibits the same general tendency with a smaller improvement. Although NSGA-II produces a substantially larger number of nondominated solutions, this should not be interpreted directly as superior Pareto quality. Its comparable HV and similar spread indicate that the larger ND count primarily reflects a denser sampling of the Pareto region, whereas the generative model produces a more compact set of nondominated alternatives. This difference is also consistent with the iterative nondominated-selection mechanism of NSGA-II, in contrast to the generative model, which produces candidate designs directly from a learned distribution. Accordingly, the combined diffusion model is used as the representative generative configuration in the subsequent budget comparison and native CAD/CAE verification. This choice reflects the empirical performance observed for the present dataset and implementation rather than a general superiority of diffusion over flow matching.

\begin{figure*}[htb!]
    \centering
    \includegraphics[width=0.95\textwidth]{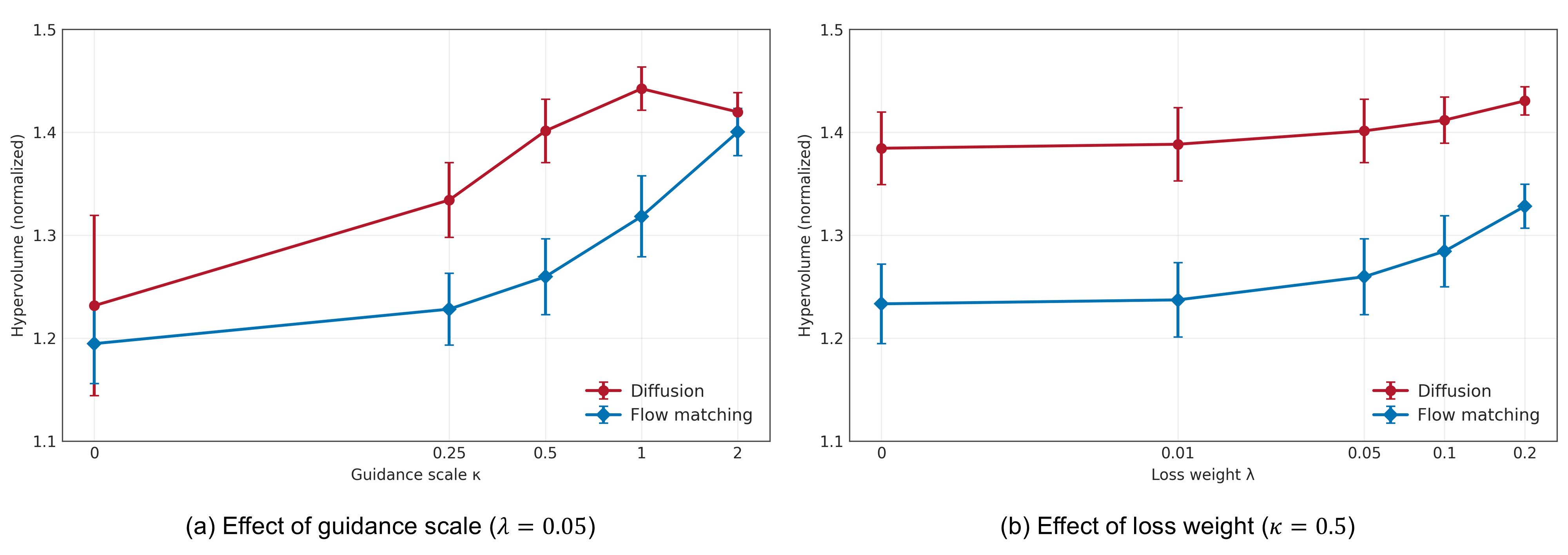}
    \caption{Sensitivity of hypervolume to the performance-incorporation parameters: (a) guidance scale $\kappa$ with $\lambda = 0.05$ and (b) performance-aware loss weight $\lambda$ with $\kappa = 0.5$.}
    \label{fig:sensitivity}
\end{figure*}

\paragraph{Sensitivity to the guidance scale and loss weight}
Based on the ablation results, the sensitivity of the combined configurations of both generative model families is further examined with respect to the performance-guidance scale $\kappa$ and the performance-aware loss weight $\lambda$ (Fig.~\ref{fig:sensitivity}). When $\lambda$ is fixed at 0.05, increasing $\kappa$ generally improves hypervolume for both diffusion and flow matching. The improvement is more pronounced for diffusion, whose hypervolume increases substantially up to $\kappa = 1.0$ and then slightly decreases at $\kappa = 2.0$, whereas flow matching exhibits a more gradual improvement. At stronger guidance levels, an increasing fraction of generated variables also approaches the prescribed design bounds, indicating a trade-off between stronger performance guidance and remaining within regions sufficiently supported by the training data. When $\kappa$ is fixed at 0.5, increasing $\lambda$ produces a more gradual improvement in hypervolume for both model families. In contrast to the guidance-scale sensitivity, flow matching shows a more pronounced relative response to increasing $\lambda$, whereas diffusion maintains a higher absolute hypervolume but changes more gradually. Overall, these results indicate that performance-guided generation and performance-aware training affect the two generative families differently, while their benefits are maintained across a range of parameter settings rather than depending on a single choice. The nominal setting of $\kappa = 0.5$ and $\lambda = 0.05$ is therefore retained for the subsequent comparisons as a moderate performance-oriented setting.

\subsection{Comparison with NSGA-II under matched evaluation budgets}
\label{subsec:nsga_comparison}
\begin{figure*}[htb!]
    \centering
    \includegraphics[width=1.00\textwidth]{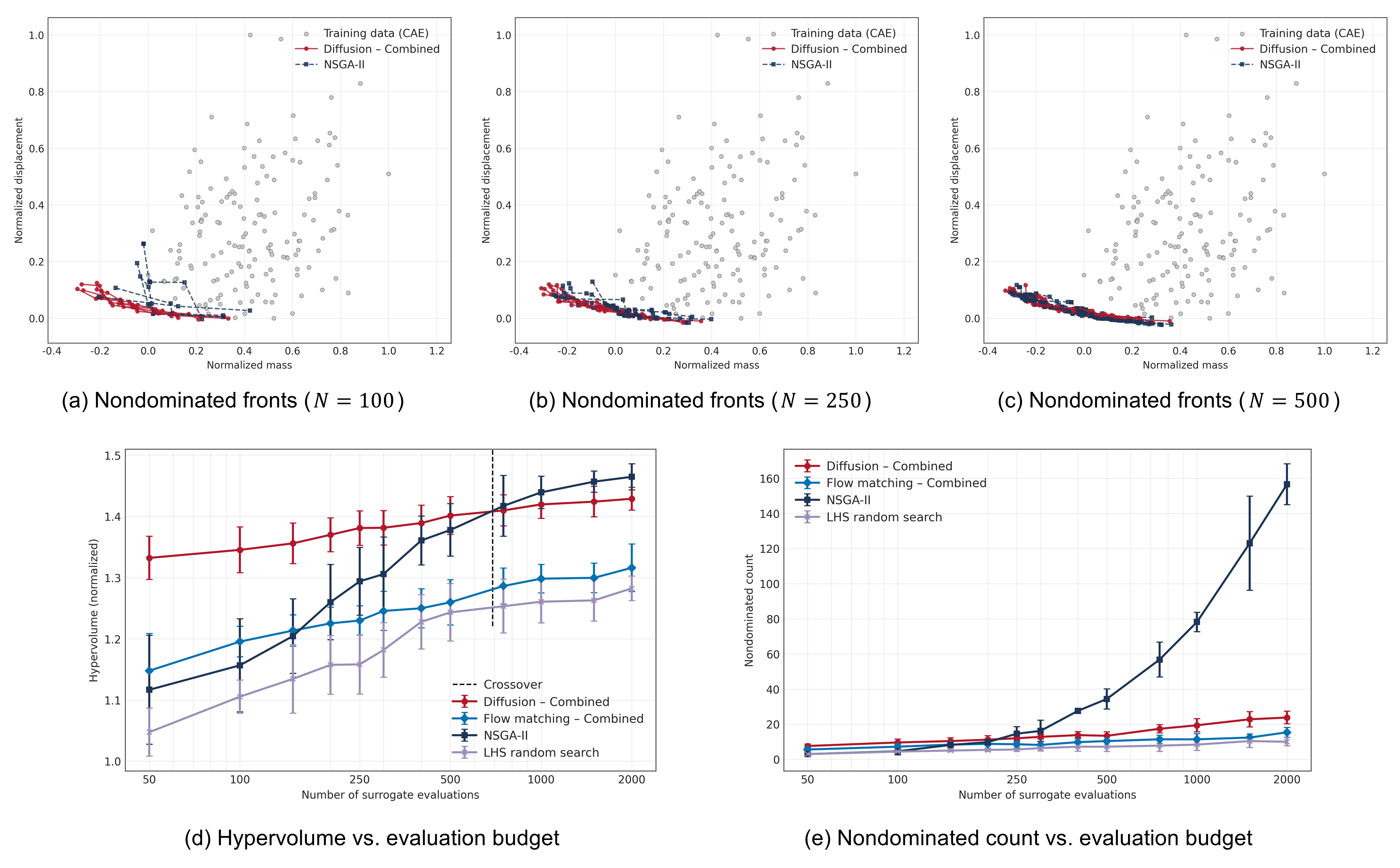}
    \caption{Budget-dependent comparison of the generative models, NSGA-II, and LHS random search. (a)--(c) Nondominated fronts of combined diffusion and NSGA-II across five seeds at $N = 100, 250, 500$, with the training data shown for reference. (d) hypervolume and (e) nondominated count as functions of the evaluation budget. The dashed line in (d) marks the approximate crossover between combined diffusion and NSGA-II.}
    \label{fig:budget_comparison}
\end{figure*}

Based on the ablation results, the combined configurations of both generative model families were included in the budget-dependent comparison, while the combined diffusion model was retained as the representative generative configuration for subsequent native CAD/CAE verification. Table~\ref{tab:budget} and Fig.~\ref{fig:budget_comparison} compare the combined diffusion and flow-matching models with NSGA-II and LHS random search over evaluation budgets ranging from 50 to 2000. Each method was evaluated over five independent seeds. For each seed, the budget-dependent results were computed cumulatively from the same sequence of evaluated candidates using the first $N$ evaluations at each budget.

\begin{table*}[htb!]
    \centering
    \caption{Hypervolume and nondominated count under matched evaluation budgets.}
    \label{tab:budget}
    
    \small
    \setlength{\tabcolsep}{3.5pt}
    \renewcommand{\arraystretch}{1.05}
    
    \begin{tabular}{@{}lcccccc@{}}
        \toprule
        \multirow{2}{*}{\textbf{Method}} 
        & \multicolumn{6}{c}{\textbf{Evaluation budget, $N$}} \\
        \cmidrule(lr){2-7}
        & 50 & 100 & 250 & 500 & 1000 & 2000 \\
        \midrule
        
        \multicolumn{7}{@{}l}{\textit{Hypervolume (normalized)}} \\
        Diffusion -- Combined
        & 1.332 $\pm$ 0.035
        & 1.345 $\pm$ 0.037
        & 1.381 $\pm$ 0.028
        & 1.401 $\pm$ 0.031
        & 1.419 $\pm$ 0.023
        & 1.429 $\pm$ 0.018 \\
        
        Flow matching -- Combined
        & 1.148 $\pm$ 0.061
        & 1.195 $\pm$ 0.025
        & 1.230 $\pm$ 0.024
        & 1.260 $\pm$ 0.037
        & 1.298 $\pm$ 0.023
        & 1.316 $\pm$ 0.039 \\
        
        NSGA-II
        & 1.117 $\pm$ 0.089
        & 1.157 $\pm$ 0.076
        & 1.294 $\pm$ 0.055
        & 1.378 $\pm$ 0.043
        & 1.439 $\pm$ 0.026
        & 1.464 $\pm$ 0.021 \\
        
        LHS random search
        & 1.047 $\pm$ 0.039
        & 1.105 $\pm$ 0.027
        & 1.158 $\pm$ 0.048
        & 1.243 $\pm$ 0.047
        & 1.260 $\pm$ 0.034
        & 1.282 $\pm$ 0.020 \\
        
        \midrule
        \multicolumn{7}{@{}l}{\textit{Nondominated count}} \\
        Diffusion -- Combined
        & 7.6 $\pm$ 1.1
        & 9.6 $\pm$ 1.9
        & 12.0 $\pm$ 2.1
        & 13.4 $\pm$ 2.2
        & 19.4 $\pm$ 3.8
        & 23.8 $\pm$ 3.6 \\
        
        Flow matching -- Combined
        & 5.6 $\pm$ 1.8
        & 7.2 $\pm$ 1.5
        & 8.6 $\pm$ 2.1
        & 10.4 $\pm$ 3.5
        & 11.4 $\pm$ 3.2
        & 15.4 $\pm$ 2.8 \\
        
        NSGA-II
        & 3.0 $\pm$ 0.7
        & 4.6 $\pm$ 2.1
        & 14.6 $\pm$ 4.0
        & 34.4 $\pm$ 5.7
        & 78.2 $\pm$ 5.5
        & 156.6 $\pm$ 11.7 \\
        
        LHS random search
        & 3.0 $\pm$ 1.2
        & 4.4 $\pm$ 0.9
        & 5.6 $\pm$ 1.1
        & 7.2 $\pm$ 2.8
        & 8.4 $\pm$ 3.4
        & 10.0 $\pm$ 2.3 \\
        
        \bottomrule
    \end{tabular}
\end{table*}

The nondominated fronts and quantitative metrics together reveal a clear budget-dependent transition between the generative and evolutionary search strategies. At $N = 100$ [Fig.~\ref{fig:budget_comparison}(a)], the combined diffusion model already concentrates solutions in the low-mass and low-displacement region, whereas the NSGA-II fronts remain relatively dispersed. At $N = 250$ [Fig.~\ref{fig:budget_comparison}(b)], NSGA-II approaches the region identified by diffusion, and by $N = 500$ [Fig.~\ref{fig:budget_comparison}(c)] the two fronts become increasingly similar. This trend is quantified by the hypervolume results in Fig.~\ref{fig:budget_comparison}(d): the combined diffusion model exceeds NSGA-II by approximately 0.19 at $N = 100$ and 0.09 at $N = 250$, the difference becomes small around $N = 500$, and the curves cross at approximately $N = 700$, after which NSGA-II gradually achieves higher Pareto quality. The combined flow-matching model shows a more gradual improvement, while LHS random search consistently exhibits lower hypervolume than the performance-oriented approaches over the tested budget range. The nondominated-count trend in Fig.~\ref{fig:budget_comparison}(e) provides a complementary perspective: NSGA-II accumulates a substantially larger number of nondominated solutions as the budget increases, whereas the generative models produce more compact nondominated sets. This larger count should not be interpreted directly as superior Pareto quality, because around $N = 500$ the combined diffusion model and NSGA-II exhibit comparable hypervolume and closely aligned fronts despite markedly different nondominated counts. The difference therefore mainly reflects denser sampling of the Pareto region by the iterative nondominated-selection process of NSGA-II, whereas the generative models directly sample candidates from a learned design distribution. This behavior is consistent with the different ways in which the two approaches exploit available information: the generative models leverage a learned prior from previously realized CAD/CAE designs and can therefore propose high-performing candidates early in the search, while NSGA-II progressively improves its solution set through adaptive evolutionary search. Consequently, the generative approach is advantageous under limited evaluation budgets, whereas the iterative refinement of NSGA-II becomes increasingly beneficial as more evaluations are allowed.

Overall, these results support a budget-dependent interpretation rather than a general superiority of either approach. A pretrained performance-oriented generator is particularly useful when high-quality alternatives must be obtained under a limited evaluation budget, whereas NSGA-II becomes more effective when sufficient evaluations are available for continued adaptive refinement. Although performance-guided generation requires additional surrogate queries during sampling, their computational cost remains small relative to that of native CAE analysis in the intended industrial workflow. The combined diffusion model is therefore carried forward to native CAD/CAE verification, with $N = 500$ retained as the working budget because it represents the transition regime in which the generative and evolutionary approaches achieve comparable Pareto quality.

\subsection{CAD/CAE verification of generated candidates}
\label{subsec:cad_cae_validation}
\begin{figure*}[htb!]
    \centering
    \includegraphics[width=0.80\textwidth]{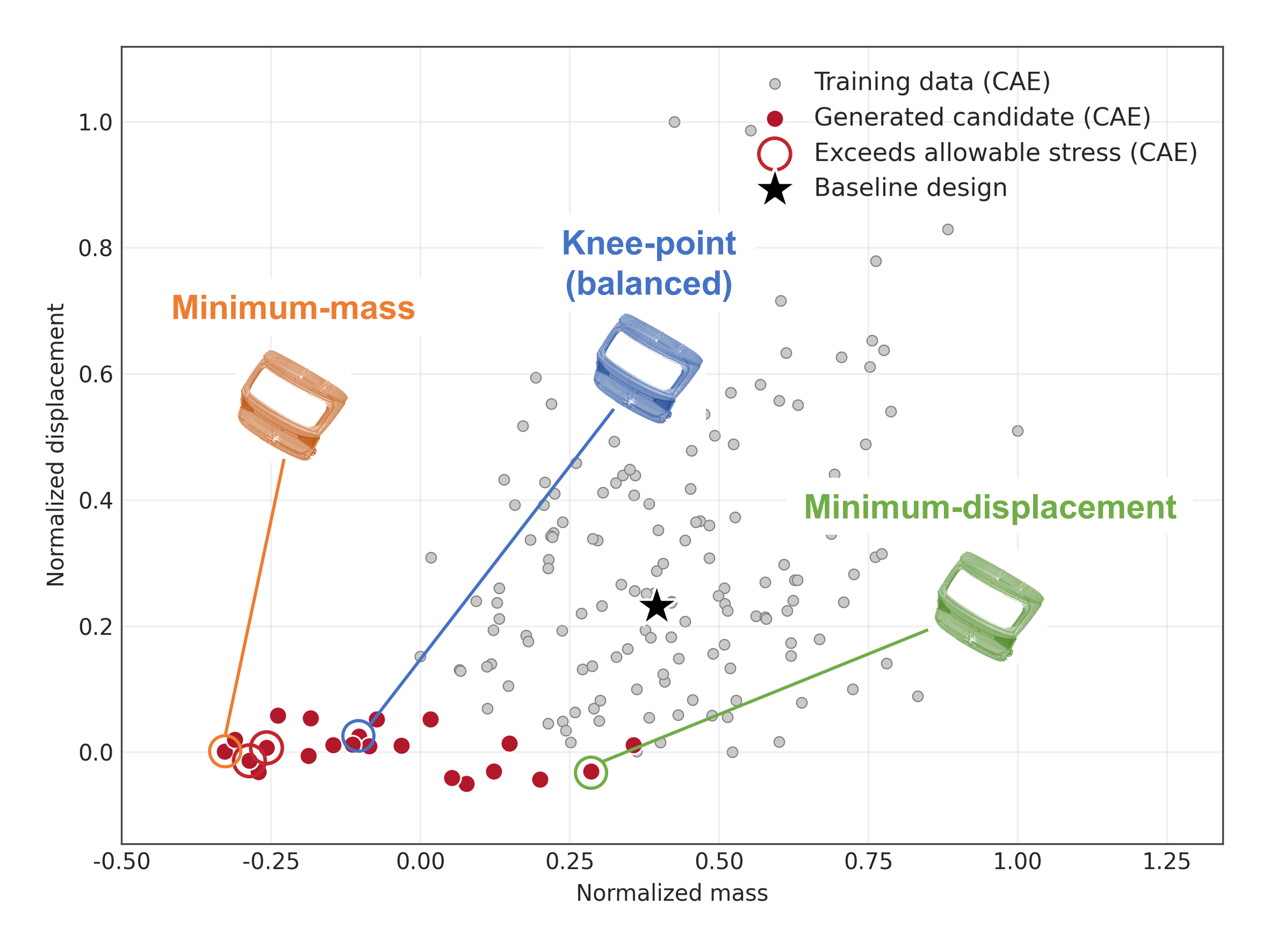}
    \caption{Native CAD/CAE verification of generated Pareto-region candidates in the normalized objective space. Training data and the baseline design are shown for reference, while circled generated candidates exceed the allowable-stress criterion. Three representative solutions are highlighted: minimum-mass, knee-point (balanced), and minimum-displacement.}
    \label{fig:direct_validation}
\end{figure*}

Direct CAD/CAE verification was performed on selected candidates distributed across the generated Pareto region. The minimum-mass, minimum-displacement, and knee-point designs were explicitly retained, while the remaining candidates were selected to provide representative coverage of the front. Here, the knee-point design was defined as the candidate closest to the utopia point after min--max normalization of the two objectives. Each selected candidate was reconstructed and re-evaluated within the native CAD/CAE workflow. Table~\ref{tab:verification} summarizes the three representative designs in detail, and Fig.~\ref{fig:direct_validation} shows the verified candidates in the objective space.

\begin{table*}[htb!]
    \centering
    \caption{Native CAD/CAE verification of the three representative designs selected from the generated Pareto front, with the baseline design listed for reference. Objective values are normalized by the min--max range of the training split. Prediction errors are computed from the original physical units. Percentage changes are calculated from the direct CAE responses relative to the baseline design.}
    \label{tab:verification}

    \small
    \setlength{\tabcolsep}{1.5pt}
    \renewcommand{\arraystretch}{1.10}

    \begin{tabular}{@{}lrrrrrrrrrc@{}}
        \toprule

        \multirow{2}{*}{\textbf{Design}}
        & \multicolumn{3}{c}{\textbf{Normalized mass}}
        & \multicolumn{3}{c}{\textbf{Normalized displacement}}
        & \multicolumn{3}{c}{\textbf{Change from baseline}}
        & \multirow{2}{*}{
            \shortstack[c]{\textbf{Structural feasibility}\\
            ($\sigma \leq \sigma_{\mathrm{allow}}$)}
        } \\
        \cmidrule(lr){2-4}
        \cmidrule(lr){5-7}
        \cmidrule(lr){8-10}

        & \multicolumn{1}{c}{Pred.}
        & \multicolumn{1}{c}{CAE}
        & \multicolumn{1}{c}{Error (\%)}
        & \multicolumn{1}{c}{Pred.}
        & \multicolumn{1}{c}{CAE}
        & \multicolumn{1}{c}{Error (\%)}
        & \multicolumn{1}{c}{$\Delta$ Mass}
        & \multicolumn{1}{c}{$\Delta$ Disp.}
        & \multicolumn{1}{c}{$\Delta$ Stress}
        & \\
        \midrule

        Baseline design
        & -- & 0.396 & --
        & -- & 0.231 & --
        & -- & -- & --
        & -- \\

        Minimum-mass
        & $-0.327$ & $-0.328$ & 0.01\%
        & 0.098 & 0.001 & 20.5\%
        & $-10.3$\%
        & $-32.8$\%
        & $-20.4$\%
        & Pass \\

        Knee-point (balanced)
        & $-0.105$ & $-0.103$ & 0.03\%
        & 0.025 & 0.025 & 0.1\%
        & $-7.1$\%
        & $-29.4$\%
        & $-17.9$\%
        & Pass \\

        Minimum-displacement
        & 0.281 & 0.286 & 0.07\%
        & $-0.015$ & $-0.030$ & 3.4\%
        & $-1.6$\%
        & $-37.3$\%
        & $-31.1$\%
        & Pass \\

        \bottomrule
    \end{tabular}
\end{table*}

\paragraph{Native CAD/CAE feasibility}
All selected candidates were successfully regenerated and analyzed in the native CAD/CAE workflow. All but two satisfied the allowable-stress criterion; the two exceptions were located near the minimum-mass end of the generated region and exceeded the allowable stress by less than 2\%, as indicated by the circled markers in Fig.~\ref{fig:direct_validation}. The successful reconstruction and analysis demonstrate that the selected Pareto-region candidates remain realizable within the native workflow. However, this result should not be interpreted as a general guarantee of feasibility, because the generative models were trained on designs that had already passed the automated CAD/CAE workflow. Direct reconstruction and CAE analysis therefore remain necessary for final engineering verification.

\paragraph{Representative designs and improvement over the baseline}
As shown in Fig.~\ref{fig:direct_validation}, the verified candidates extend beyond the training data toward the low-mass and low-displacement region, while the baseline design lies within the original training distribution. The three highlighted designs represent the two objective extremes and a balanced trade-off along the generated front. Table~\ref{tab:verification} shows that all three satisfy the allowable-stress criterion and achieve substantial improvements after direct CAE verification. Relative to the baseline, the knee-point design reduced mass by 7.1\% and maximum displacement by 29.4\%, together with a 17.9\% reduction in maximum von Mises stress. The minimum-mass design achieves the largest mass reduction of 10.3\%, whereas the minimum-displacement design achieves the largest displacement reduction of 37.3\%. All three representative designs simultaneously reduce mass, displacement, and stress relative to the baseline.

The surrogate predictions for mass remain highly consistent with the direct CAE results for all three representative designs, while larger discrepancies are observed for displacement, particularly toward the minimum-mass end of the front. Nevertheless, the CAE responses of all three designs remain within the 95\% predictive intervals of the surrogate. These results support the intended two-stage use of the framework: surrogate-assisted generative search identifies a promising Pareto region, while native CAD reconstruction and CAE analysis provide final verification of selected design candidates.

\subsection{Engineering implications}
\label{subsec:engineering implications}
The case study highlights that industrial generative design optimization depends not only on the performance of the learning models but also on how engineering knowledge and feasibility are represented throughout the workflow. Reference-based CAD preprocessing defines a controlled design space in which generated parameters can be mapped back to an editable product model, while explicit CAD/CAE failure handling distinguishes nominal parameter feasibility from feasibility in the native engineering environment. The results further indicate that performance-oriented generative search should be interpreted together with surrogate reliability, evaluation budget, and native CAD/CAE verification. These results suggest that the proposed framework is best viewed as a hybrid engineering--AI workflow in which domain modeling, data-driven exploration, and engineering verification jointly support design decision making.

\section{Conclusions and future work}
\label{sec:conclusions_future_work}
This study presented a generative AI-driven parametric design optimization framework for industrial vehicle closure systems. The central contribution of the framework is to connect performance-oriented generative search with an editable industrial CAD assembly and native CAD/CAE verification within a unified engineering workflow. Engineering performance is incorporated into the generative process through performance-guided generation with surrogate guidance and performance-aware training using a performance-aware loss, enabling diffusion and flow matching models to direct design exploration toward high-performing regions of the learned design space. The generated design variables are then mapped back to a reference-based CATIA assembly, regenerated in the native CAD environment, and re-evaluated through CAE so that predicted performance improvements can be assessed together with CAD/CAE feasibility. The automotive tailgate case study therefore demonstrates a practical bridge between performance-oriented generative design exploration and detailed product engineering, while retaining native CAD editability and downstream engineering verification.

The results showed that the automated workflow produced 189 valid design--performance pairs from 200 sampled designs while explicitly identifying CAD- and CAE-level failures within the prescribed design space. The developed GPR surrogate achieved a hold-out \(R^2\) of 0.9998 for mass and 0.8889 for maximum displacement, with well-calibrated PIs, providing the performance estimates required for subsequent design exploration. Performance-guided generation using surrogate information shifted generated candidates toward improved performance regions, while performance-aware training further improved the stability and quality of the generated design distribution; the combined strategy provided the best Pareto performance for both generative families and was robust to the guidance scale and loss weight. Under matched evaluation budgets, the comparison with NSGA-II revealed a budget-dependent crossover: the generative model achieved higher Pareto quality under small candidate-evaluation budgets, the two methods were comparable at 500 evaluations, and NSGA-II became stronger beyond approximately 700 evaluations. Direct CAD reconstruction and CAE re-evaluation showed that all selected candidates were successfully regenerated and analyzed, with all but two satisfying the allowable-stress criterion. The knee-point design that balances both objectives reduced mass and maximum displacement by 7.1\% and 29.4\%, respectively, relative to the baseline design, with reductions of up to 10.3\% in mass and 37.3\% in maximum displacement at the two extremes of the verified Pareto front.

From an engineering design perspective, these results highlight that numerical performance improvement alone is insufficient for industrial generative design optimization. Candidate designs must also remain compatible with the parametric dependencies, geometric regeneration requirements, and downstream analysis procedures embedded in the product-development workflow. The present study is limited to a single tailgate architecture and design skin, a primarily structural design problem, and a relatively limited CAD/CAE dataset, which may affect surrogate reliability when generated candidates approach sparsely sampled regions of the design space. In addition, the current generative models operate at the parameter level rather than at the feature or functional level. Future work will extend the framework to multiple design skins and broader multidisciplinary performance requirements, incorporate CAD/CAE feasibility directly into the generative process, and account explicitly for surrogate uncertainty during performance guidance. A further direction is to move from parameter-level generation toward feature- and function-level design generation so that higher-level designer intent can be incorporated while retaining native CAD editability. These developments would support the transition from task-specific parametric design exploration toward more reusable AI-assisted engineering design systems.

\section*{CRediT authorship contribution statement}
\textbf{Dongeon Lee}: Conceptualization, Methodology, Software, Formal analysis, Investigation, Validation, Visualization, Writing -- original draft.
\textbf{Yurim Lee}: Methodology, Software, Data curation, Investigation, Validation, Visualization, Writing -- original draft.
\textbf{Jiyong Kim}: Methodology, Software, Data curation, Investigation, Validation.
\textbf{Sanghyeon Kim}: Methodology, Software, Data curation, Investigation, Validation.
\textbf{Sanghyun Lee}: Resources, Project administration, Writing -- review \& editing.
\textbf{Seungyeon Shin}: Resources, Project administration, Writing -- review \& editing.
\textbf{Jeahong Kim}: Resources, Project administration, Writing -- review \& editing.
\textbf{Namwoo Kang}: Supervision, Conceptualization, Funding acquisition, Writing -- review \& editing.


\section*{Declaration of competing interest}
The authors declare that they have no known competing financial interests or personal relationships that could have appeared to influence the work reported in this paper.

\section*{Data availability}
The industrial CAD geometry, raw design-variable definitions, detailed CAE settings, and full design-performance dataset are not publicly available because they contain proprietary information of Hyundai Motor Company. Derived or normalized data may be made available by the corresponding author upon reasonable request, subject to approval under the industry--academic collaboration agreement.

\section*{Acknowledgments}
This work was supported by Hyundai Motor Company and by grants from the Ministry of Science and ICT (GTL24033-000, N1026002, and No. 2022-0-00986), the Ministry of Trade, Industry and Energy (RS-2025-02317327 and RS-2025-25444634), the Ministry of Oceans and Fisheries (PET0050), and Korea Hydro \& Nuclear Power Co., Ltd. (No. 8-Tech-07).

\bibliographystyle{unsrtnat}
\bibliography{references}

\end{document}